\pdfoutput=1

\documentclass[11pt]{article}

\usepackage[]{acl}

\usepackage{times}
\usepackage{latexsym}
\usepackage{tabularx}
\usepackage{booktabs}
\usepackage{amsmath}
\usepackage{graphicx}
\usepackage{enumitem}
\usepackage{boxedminipage}

\usepackage[T1]{fontenc}
\usepackage{amsfonts}

\usepackage[utf8]{inputenc}

\usepackage{microtype}
\DeclareMathOperator*{\argmin}{arg\,min}
\definecolor{darkgreen}{rgb}{0.0, 0.5, 0.0}
\everypar{\looseness=-1}
\title{\textit{IsoVec}: Controlling the Relative Isomorphism of Word Embedding Spaces}

\author{Kelly Marchisio, Neha Verma, Kevin Duh, \and Philipp Koehn \\
 Johns Hopkins University \\ \tt{ \{kmarc, nverma7\}@jhu.edu, kevinduh@cs.jhu.edu, phi@jhu.edu}}

\begin{document}
\maketitle
\begin{abstract}
The ability to extract high-quality translation dictionaries from monolingual word embedding spaces depends critically on the geometric similarity of the spaces---their degree of ``isomorphism.''  We address the root-cause of faulty cross-lingual mapping: that word embedding training resulted in the underlying spaces being non-isomorphic. 
We incorporate global measures of isomorphism directly into the Skip-gram loss function, successfully increasing the relative isomorphism of trained word embedding spaces and improving their ability to be mapped to a shared cross-lingual space. The result is improved bilingual lexicon induction in general data conditions, under domain mismatch, and with training algorithm dissimilarities.  
We release \textit{IsoVec} at \url{https://github.com/kellymarchisio/isovec}.
\end{abstract}

\section{Introduction}

\begin{figure*}[]
\centering
\includegraphics[height=0.25\textheight,width=0.95\linewidth]{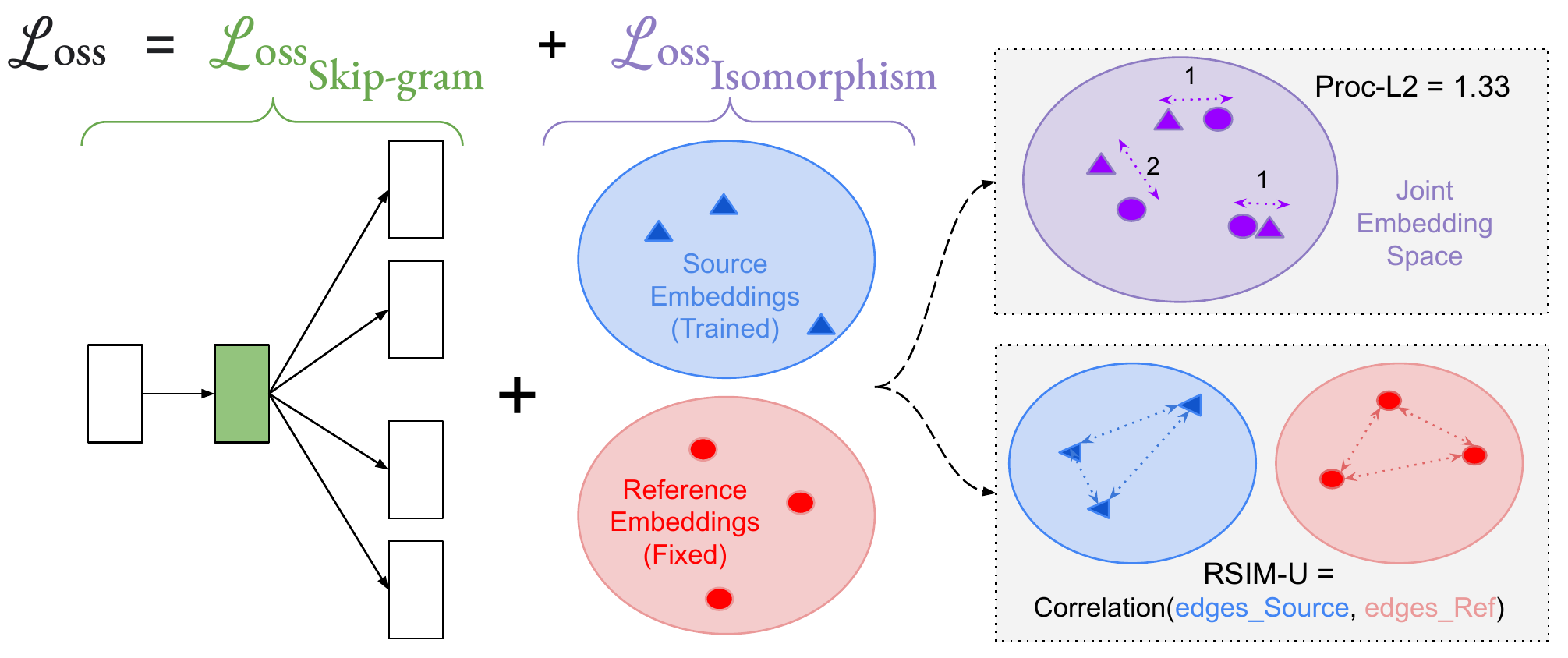}
\caption{Proposed Method. Loss is a weighted combination of Skip-gram with negative sampling loss (seen left with a reproduction of the familiar image from \citet{mikolov2013efficient} for reader recognizability) and an isomorphism loss (seen right, ours) calculated in relation to a fixed reference space.  Gray boxes are two possibilities explored in this work: Proc-L2 (supervised) where $L_{ISO}$ is calculated over given seed translations, and RSIM-U (unsupervised).}
\label{fig:isovec}
\end{figure*}

The task of extracting a translation dictionary from word embedding spaces, called ``bilingual lexicon induction'' (BLI), is a common task in the natural language processing literature. 
Bilingual dictionaries are useful in their own right as linguistic resources, and automatically generated dictionaries may be particularly helpful for low-resource languages for which human-curated dictionaries are unavailable. 
BLI is also used as an extrinsic evaluation task to assess the quality of cross-lingual spaces.  
If a high-quality translation dictionary can be automatically extracted from a shared embedding space, intuition says that the space is high-quality and useful for downstream tasks.  

``Mapping-based'' methods are one way to create cross-lingual embedding spaces. 
Separately-trained monolingual embeddings are mapped to a shared space by applying a linear transformation to one or both spaces, after which a bilingual lexicon can be extracted via nearest-neighbor search
\cite[e.g.,][]{mikolov2013, lample2018word, artetxe-etal-2018-robust,  joulin-etal-2018-loss,patra-etal-2019-bilingual}.

Mapping methods are effective for closely-related languages with embedding spaces trained on high-quality, domain-matched data even without supervision, but critically rely on the ``approximate isomorphism assumption''---that monolingual embedding spaces are geometrically similar.\footnote{In formal mathematicals, “isomorphic” requires two objects to have an invertible correspondence between them. Researchers in NLP loosen the definition to ``geometrically similar'', and consider \textit{degrees} of similarity. 
We might say that space X is \textit{more isomorphic} to space Y than is space Z.}  
Problematically, researchers have observed that the isomorphism assumption weakens substantially as languages and domains become dissimilar, leading to failure precisely where unsupervised methods might be helpful \cite[e.g.][]{sogaard-etal-2018-limitations,ormazabal-etal-2019-analyzing,glavas-etal-2019-properly, vulic-etal-2019-really,patra-etal-2019-bilingual,marchisio-etal-2020-unsupervised}.  

Existing work attributes non-isomorphism to linguistic, algorithmic, data size, or domain differences in training data for source and target languages. 
From \citet{sogaard-etal-2018-limitations}, ``the performance of unsupervised BDI [BLI] depends heavily on... language pair, the comparability
of the monolingual corpora, and the parameters of the word embedding algorithms.'' Several authors found that unsupervised machine translation methods suffer under similar data shifts \cite{marchisio-etal-2020-unsupervised, kim2020and,marie-unsup-mt-2020}.

While such factors do result in low isomorphism of spaces trained with traditional methods, we needn't resign ourselves to the mercy of the geometry a training methodology naturally produces. 
While multiple works post-process embeddings or map non-linearly, we control similarity explicitly during embedding training by incorporating five global metrics of isomorphism into the Skip-gram loss function.
Our three supervised and two unsupervised losses gain some control of the relative isomorphism of word embedding spaces, compensating for data mismatch and creating spaces that are linearly mappable where previous methods failed.

\section{Related Work}
\paragraph{Cross-Lingual Word Embeddings}
There is a broad literature on creating cross-lingual word embedding spaces. Two major paradigms are ``mapping-based'' methods which find a linear transformation to map monolingual embedding spaces to a shared space \cite[e.g.,][]{artetxe-etal-2016-learning, artetxe-etal-2017-learning,alvarez-melis-jaakkola-2018-gromov,doval-etal-2018-improving, jawanpuria-etal-2019-learning}, 
and ``joint-training'' which, as stated in the enlightening survey by \citet{ruder2019survey}, ``minimize the source and target language monolingual losses jointly with the cross-lingual regularization term'' \cite[e.g.][\citet{ruder2019survey} for a review]{luong-etal-2015-bilingual}. \citet{gouws-2015} train Skip-gram for source and target languages simultaneously, enforcing an L2 loss for known translation. \citet{Wang*2020Cross-lingual} compare and combine joint and mapping approaches.

More recently, researchers have explored massively multilingual language models \cite{devlin-etal-2019-bert, conneau-etal-2020-unsupervised}. 
While these have been shown to possess some inherent cross-lingual transfer ability \cite{wu-dredze-2019-beto}, another line of work focuses on improving their cross-lingual representations with explicit cross-lingual signal \cite{wang-etal-2019-cross, liu-etal-2019-investigating, Cao2020Multilingual, kulshreshtha-etal-2020-cross, wu-dredze-2020-explicit}. Recently, \citet{li-etal-2022-improving} combined static and pretrained multilingual embeddings for BLI.

\paragraph{Handling Non-Isomorphism}
\citet{miceli-barone-2016-towards} explore whether comparable corpora induce embedding spaces which are approximately isomorphic. 
\citet{ormazabal-etal-2019-analyzing} compare cross-lingual word embeddings induced via mapping methods 
and jointly-trained embeddings from \citet{luong-etal-2015-bilingual}, finding that the latter are better in measures of isomorphism and BLI precision.
\citet{nakashole-flauger-2018-characterizing} argue that word embedding spaces are not globally linearly-mappable.  Others use non-linear mappings \cite[e.g.][]{mohiuddin-etal-2020-lnmap, glavas-vulic-2020-non} or post-process embeddings after training to improve quality \citep[e.g.][]{peng-etal-2021-cross, faruqui-etal-2015-retrofitting, mu2018allbutthetop}. \citet{eder-etal-2021-anchor} initialize a target embedding space with vectors from a higher-resource source space, then train the low-resource target.
 \citet{zhang-etal-2017-earth} minimize earth mover's distance over 50-dimensional pretrained word2vec embeddings. \citet{ormazabal-etal-2021-beyond} learn source embeddings in reference to fixed target embeddings given known or hypothesized translation pairs induced during via self-learning.

\paragraph{Examining \& Exploiting Embedding Geometry} Emerging literature examines geometric properties of embedding spaces. In addition to isomorphism, some examine \textit{isotropy} \cite[e.g.][]{mimno-thompson-2017-strange,mu2018allbutthetop,ethayarajh-2019-contextual, rajaee-pilehvar-2022-isotropy, rudman-etal-2022-isoscore}.
\citet{li-etal-2020-sentence} transform the semantic space from masked language models into an isotropic Gaussian distribution from a non-smooth anisotropic space. \citet{bert_whitening} apply whitening and dimensionality reduction to improve isotropy. 
\citet{zhang-etal-2022-effect} inject isotropy into a variational autoencoder, and \citet{ethayarajh-jurafsky-2021-attention} recommend ``adding an anisotropy penalty to the language modelling objective'' as future work. 

\section{Background}
We discuss the mathematical background used in our methods. 
Throughout, $\bf{X} \in \mathbb{R}^{n \times d}$ and $\bf{Y} \in \mathbb{R}^{m \times d}$ are the source and target word embedding spaces of $d$-dimensional word vectors, respectively.  We may assume seed pairs $\{(x_0, y_0), (x_1, y_1),... (x_s, y_s)\}$ are given.

\subsection{The Orthogonal Procrustes Problem}\label{sec:proc}
\citet{schonemann1966generalized} derived the solution to the orthogonal Procrustes problem, whose goal is to find the linear transformation $W$ that solves: 
\begin{equation*}
    \argmin_{W \in \mathbb{R}^{d \times d}, W^T W=I} ||X W-Y||_F^2 
\label{eq:proc-eq}
\end{equation*}
The solution is $W = VU^T$, where $U\Sigma V^T$ is the singular value decomposition of $Y^TX$. If $X$ is a matrix of vectors corresponding to seed words $x_i$ in $\{(x_0, y_0), (x_1, y_1), \ldots, (x_s, y_s)\}$ and $Y$ is a matrix of the corresponding $y_i$, then $W$ is the linear transformation that minimizes the difference between the vector representations of known pairs. 

\subsection{Embedding Space Mapping with VecMap}
We use the popular VecMap\footnote{\url{https://github.com/artetxem/vecmap}} toolkit for embedding space mapping, which can be run in supervised, semi-supervised, and unsupervised modes. As of the time of its writing, \citet{glavas-etal-2019-properly} deem VecMap the most robust unsupervised method. 

First, source and target word embeddings are unit-normed, mean-centered, and unit-normed again \cite{zhang-etal-2019-girls}. The bilingual lexicon is induced by whitening each space and then solving a variant of the orthogonal Procrustes problem.\footnote{See Appendix \ref{sec:app-white}, \ref{sec:app-ortho} for details} Spaces are reweighted and dewhitened, and translation pairs are extracted via nearest-neighbor search from the mapped embedding spaces. See the original works and implementation for details \cite{artetxe2018generalizing, artetxe-etal-2018-robust}.

Unsupervised and semi-supervised modes utilize the same framework as supervised mode, but with an iterative self-learning procedure that repeatedly solves the orthogonal Procrustes problem over hypothesized translations. On each iteration, new hypotheses are extracted. 
The modes differ only in how they induce the initial hypothesis seed pairs.   
In semi-supervised mode, this is a given input seed dictionary. In unsupervised mode, similarity matrices $M_x=XX^T$ and $M_z=ZZ^T$ are created over the first $n$ vocabulary words.\footnote{Default: 4000} Word $z_j$ is the assumed translation of $x_i$ if vector $M_{z_j}$ is most similar to $M_{x_i}$ compared to all others in $M_z$.   See \citet{artetxe-etal-2018-robust} for details. 

\subsection{Isomorphism Metrics}
In NLP, relative isomorphism is often measured by Relational Similarity, Eigenvector Similarity, and Gromov-Hausdorff Distance. We describe these metrics in detail in this section.

\paragraph{Relational Similarity} Given seed translation pairs, calculate pairwise cosine similarities:
\begin{footnotesize}
\begin{gather*}
\text{cos}(x_0, x_1) \qquad \text{cos}(y_0, y_1) \\
\text{cos}(x_0, x_2) \qquad \text{cos}(y_0, y_2) \\
\text{cos}(x_0, x_3) \qquad \text{cos}(y_0, y_3) \\
\dots \qquad \qquad \dots \\
\text{cos}(x_1, x_0) \qquad \text{cos}(y_1, y_0) \\
\text{cos}(x_1, x_2) \qquad \text{cos}(y_1, y_2) \\
\dots \qquad \qquad \dots \\
\text{cos}(x_s, x_s) \qquad \text{cos}(y_s, y_s) 
\end{gather*}
\end{footnotesize}
The Pearson's correlation between the lists of cosine similarities is known as Relational Similarity \cite{vulic-etal-2020-good, zhang-etal-2019-girls}. 

\paragraph{Eigenvector Similarity} \cite{sogaard-etal-2018-limitations} measures isomorphism between two spaces based on the Laplacian spectra of their $k$-nearest neighbor ($k$-NN) graphs. 
For seeds $\{x_0, x_1, \hdots, x_s\}$ and $\{y_0, y_1, \hdots, y_s\}$, we compute unweighted $k$-NN graphs $G_X$ and $G_Y$, then compute the Graph Laplacians ($L_G$) for both graphs (the degree matrix minus the adjacency matrix: $L_G = D_G - A_G$). We then compute the eigenvalues of $L_{G_X}$ and $L_{G_Y}$, namely $\{\lambda_{L_{G_X}}(i)\}$ and $\{\lambda_{L_{G_Y}}(i)\}$. We select $l = \min(l_X, l_Y)$ where $l_X$ is the maximum $l$ such that the first $l$ eigenvalues of $L_{G_X}$ sum to less than 90\% of the total sum of the eigenvalues. EVS is the sum of squared differences between the partial spectra:
\begin{equation*}
    \text{EVS} = \sum_{i=1}^l (\lambda_{L_{G_X}}(i) - \lambda_{L_{G_Y}}(i))^2 
\end{equation*}

The Laplacian allows one to decompose a graph function into a sum of weighted Laplacian eigenvectors, which roughly corresponds to a frequency-based decomposition of the function. The weights (eigenvalues) determine the contribution of each eigenvector to the final function. Graphs with similar Laplacian eigenvalues should have similar structure, which is what EVS aims to capture. 

\begin{figure}[]
\centering
\includegraphics[height=0.12\textheight,width=1\linewidth]{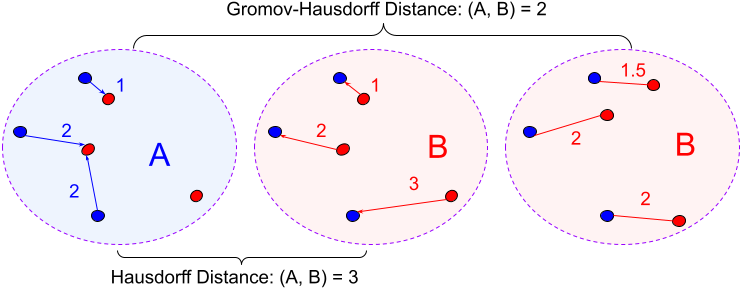}
\caption{Calculation of Gromov-Hausdorff (GH) Distance: the worst case distance of nearest neighbors in a shared embedding space after optimal orthogonal mapping. The right-most red dots have been orthogonally rotated to turn Hausdorff distance into GH Distance.}
\label{fig:gh}
\end{figure}

\paragraph{Gromov-Hausdorff Distance} is a “worst-case” metric that optimally linearly maps embedding spaces and then calculates the distance between nearest neighbors in a shared space.
\begin{itemize}[]
    \item For each $x$ of the source embeddings, find its nearest neighbor $y$ of the target embeddings.  Measure the distance. 
    \item For each $y$ of the target embeddings, find its nearest neighbor $x$ of the source embeddings.  Measure the distance.
    \item \textit{Hausdorff distance} is the worst of the above.
    \item \textit{Gromov-Hausdorff distance} is Hausdorff distance after optimal isometric transformation to minimize distances. As in previous work, since we apply mean-centering to source and target embeddings, we search only over the space of orthogonal transformations \cite{patra-etal-2019-bilingual, vulic-etal-2020-good}. See Figure \ref{fig:gh}.
\end{itemize}

\noindent We follow \citet{chazal-2009-bottleneck} and approximate the Gromov-Hausdorff distance with the Bottleneck distance between the source and target embeddings.

\section{Method}
We implement Skip-gram with negative sampling on GPU using PyTorch and use it to train monolingual embedding spaces for Bengali (bn), Ukranian (uk), Tamil (ta), and English (en).\footnote{Dim: 300, window: 5, negative samples: 10, min\_count: 10, batch size: 16384, LR: 0.001.  Adam with linear warmup for 1/4 of batches, then polynomial decay.  Run 10 epochs.} Our implementation mirrors the official word2vec\footnote{\url{https://github.com/tmikolov/word2vec}} release closely \cite{mikolov2013efficient}.

We create comparison embedding spaces using the official word2vec release with default hyperparameters and map the resulting spaces from both algorithms with VecMap for BLI. We report precision@1 (P@1) on the development set in Table \ref{tab:baseline}. P@1 is a standard evaluation metric for BLI. Our implementation slightly outperforms word2vec except ta in unsupervised mode.

\begin{table}[htb]
\footnotesize
\begin{center}
\setlength{\tabcolsep}{2.7pt}
\begin{tabular}{@{}c|c|c|c|c||c|c|c||c|c|c@{}}
\toprule
& & \multicolumn{3}{c||}{\bf{bn-en}} & \multicolumn{3}{c||}{\bf{uk-en}}  & \multicolumn{3}{c}{\bf{ta-en}}\\
\hline
& & \bf Su & \bf Se & \bf U & \bf Su & \bf Se & \bf U & \bf Su & \bf Se & \bf U \\
\hline
W2V & $\mu$ & 12.6 & 11.9 & 9.0 & 10.8 & 9.7 & 7.7 & 8.4 & 7.3 & 7.1 \\
& $\sigma$ & \itshape 0.3 & \itshape 0.3 & \itshape 5.0 & \itshape 0.5	& \itshape 0.4 & \itshape 4.3 & \itshape 0.5 & \itshape 0.6 & \itshape 0.6 \\
\hline
\hline
Ours & $\mu$  & 13.1 & 12.2 & 10.8 & 12.4 & 11.7 & 10.5 & 9.3 & 8.3 & 1.8 \\
& $\sigma$ & \itshape 1.0 & \itshape 0.4 & \itshape 0.9 & \itshape 0.9 & \itshape 0.5 & \itshape 0.6 & \itshape 0.7 & \itshape 0.9 & \itshape 3.7 \\
\hline
\end{tabular}
\end{center}
\caption{\label{tab:baseline} P@1 of mapped embedding spaces on the development set: our implementation vs. official word2vec release. Shown: Mean ($\mu$) and standard deviation ($\sigma$) over 5 runs in supervised (Su), semi-supervised (Su), and unsupervised (U) modes.} 
\end{table}

\begin{table}[htb]
\footnotesize
\begin{center}
\setlength{\tabcolsep}{2.8pt}
\begin{tabular}{@{}ccccc|c|c@{}}
\toprule
& \multicolumn{4}{c|}{{\underline{newscrawl2020}}} &  {\underline{Common Crawl}} & {\underline{newscrawl2018-20}} \\
& \textbf{en} & \textbf{bn} & \textbf{ta} & \textbf{uk} & \textbf{en} & \textbf{en} \\
\midrule
& 29.0 & 14.7 & 12.6 & 7.8 & 750 & 2700 \\
\bottomrule
\end{tabular}
\end{center}
\caption{\label{tab:datasizes} Size of training data (millions of tokens).} 
\end{table}

\subsection{Data}
For the main experiments, we train word embeddings on the first 1 million lines from newscrawl2020 for en, bn, and ta \cite{barrault-etal-2020-findings}.\footnote{\url{https://data.statmt.org/news-crawl/}} For uk, we use the entirety of newscrawl2020 ($\sim$427,000 lines). We normalize punctuation, lowercase, remove non-printing characters, and tokenize using standard Moses scripts.\footnote{\url{github.com/moses-smt/mosesdecoder/tree/master/scripts/tokenizer}} Domain mismatch experiments in Section \ref{sec:dom-mismatch} use approximately 33.8 million lines of webcrawl from the English Common Crawl.  Larger data experiments in the same section use 93 million lines of English newscrawl2018-2020. The size of the training data in tokens is seen in Table \ref{tab:datasizes}.
 
We use the publicly available train and test dictionaries from MUSE \cite{lample2018word}.\footnote{\url{https://github.com/facebookresearch/MUSE\#ground-truth-bilingual-dictionaries}} For the development set, we use source words 6501-8000 from the ``full'' set. Train, development, and test sets are non-overlapping. We use all possible training set seed words for our supervised losses, which is 6000-7000 word pairs per language.\footnote{$\sim90\%$ of train set pairs are present in the trained embedding spaces; bn-en: 6859, uk-en: 6476, ta-en: 6019.}  We use the test set for evaluating downstream BLI.

\subsection{Integrating Isomorphism Losses}
\label{sec:iso-metrics}
To train the embedding space $\bf{X}$ such that it 1) captures the distributional information via Skip-gram with negative sampling and 2) is geometrically similar to the reference word embedding space $\bf{Y}$, we propose the objective below. 
$\mathcal{L}_{SG}$ is the familiar Skip-gram with negative sampling loss function and $\mathcal{L}_{ISO}$ is the isomorphism metric loss. 
Each $\mathcal{L}_{ISO}$ requires a reference embedding space $\bf{Y}$, trained separately using our base implementation. 
\begin{equation*}
    \mathcal{L} = (1 - \beta) \, \mathcal{L}_{SG} + \beta\, \mathcal{L}_{ISO}
\end{equation*}
We use English as the reference language because we generally assume that the data quality is higher than the low-resource languages used on the source-side. 
$\bf{Y}$ is normalized, mean-centered, and normalized again before use.  
On each calculation of $\mathcal{L}_{ISO}$, we perform the same operations on a copy of the current model's word embeddings.  

\subsection{Supervised Losses}
\label{sec:sup_losses}
We assume seeds $\{(x_0, y_0), (x_1, y_1),... (x_s, y_s)\}$.  $X$ is a matrix of the current model's word embeddings for [$x_0, x_1, ..., x_s$] and $Y$ is the matrix of reference source embeddings for [$y_0, y_1, ..., y_s$].
\paragraph{L2}  
We implement L2 distance and normalize over samples. Intuitively, this coaxes translation pairs to have similar vector representations, with the hope that other words in $\bf{X}$ and $\bf{Y}$ will be tugged closer to their translations. \textbf{L2} is easy to implement and understand, and computes quickly.     
\[ \frac{1}{|X|}\sum_{i=1}^{s} || x_i - y_i ||_2 \]
\paragraph{Proc-L2} We find $W$ that solves the orthogonal Procrustes problem as in Section \ref{sec:proc}, then minimize L2 distance over the mapped space:
\[ \frac{1}{|X|} \sum_{i=1}^{s} || x_iW - y_i ||_2 \]
\paragraph{Proc-L2+Init}
Same as \textbf{Proc-L2}, except initialize source seed embeddings with the reference translation vectors so that spaces begin with the same representation for known translations.

\paragraph{RSIM} We implement relational similarity over seeds. Higher is better, so we minimize $\mathcal{L}_{ISO} = 1-\text{Pearsons\_Corr}$. 
Like \textbf{Proc-L2+Init}, we can also initialize the source space with reference seed embeddings. We call this \textbf{RSIM+Init}.

\subsection{Unsupervised Losses}
\label{sec:unsup_losses}
We use two unsupervised metrics to increase isomorphism when no seed translations are available. 

\paragraph{RSIM-U} In this unsupervised variant of \textbf{RSIM}, we calculate pairwise cosine similarities over the first $k$ words in $\bf X$ and $\bf Y$, sort the lists, then calculate Pearson's correlation. As above, $\mathcal{L}_{ISO} = 1-\text{Pearsons\_Corr}$.  We use $k=2000$ for efficiency. 

\paragraph{EVS-U} We calculate eigenvector similarity over the first 2000 words in $\bf X$ and $\bf Y$.

\subsection{On Differentiability}
Each metric must be differentiable with respect to $\bf X$, a matrix of the model's current word embeddings, to allow isomorphism-based losses to inform parameter updates in $\bf X$. 

\textbf{L2} is straightforwardly differentiable, as it is the Frobenius norm of $X-Y$. The same applies for variants \textbf{Proc-L2} and \textbf{Proc-L2+Init}. \textbf{RSIM} is naturally differentiable, seen in the formulation below. For mean-centered cosine similarity vectors\footnote{Mean-centering and cosine similarity are differentiable.} $x_{sim}$ and $y_{sim}$, Pearson's correlation coefficient is:
\begin{align*}
    \frac{\sum_i x_{sim}(i) y_{sim}(i)}{\sqrt{\sum_i x_{sim}(i)^2 y_{sim}(i)^2}} = \frac{x_{sim}^Ty_{sim}}{\lVert x_{sim}\rVert \lVert y_{sim}\rVert}
\end{align*}

\textbf{EVS} is not immediately differentiable due to the need for the non-differentiable $k$-NN computation. Instead we modify the graph computation step to use a fully-connected \textit{weighted} graph where the edge weight is the dot product between node vectors.\footnote{$x_i ^T y_j\le 1$, as all vectors are unit-normalized.} With this amended formulation, computing the gradients of Laplacian eigenvalues is possible.

\subsection{$\beta$ and Linear Mapping for BLI}
\label{sec:beta}
Each isomorphism loss may be considered a different method, as each loss may cause the overall framework to behave differently. 
Accordingly, we set $\beta$ for each separate loss function based on performance on the development set.\footnote{We try $\beta\in\{0.5, 0.333, 0.2, 0.1, 0.01\}$. For RSIM* and EVS-U, we also try $0.001$. An early L2 run used 0.05, 0.0001.}    
After selecting $\beta$, we evaluate and present results only on the test set.
$\beta$s for each method are in Table \ref{tab:beta}.  
\begin{table}[htb]
\centering  \setlength\tabcolsep{3.5pt}\footnotesize
\begin{tabular}{l|c|cl|cll|c}
\toprule
& L2 & Proc-L2 & +Init & RSIM & +Init & -U & EVS-U \\
\midrule 
$\beta$ & 0.1 & 0.333 & 0.2 & 0.01 & 0.001 & 0.1 & 0.333 \\
\bottomrule
\end{tabular}
\caption{$\beta$ parameter for isomorphism losses. Each loss function should be considered a separate method, so $\beta$ is set for each loss based on development set performance.  Once $\beta$ is chosen, we evaluate on the test set.}
\label{tab:beta}
\end{table}

VecMap in supervised mode consistently scores higher than semi-supervised mode in all baseline experiments on the development set.  
For \textit{IsoVec}, semi-supervised mapping often works best. We thus use VecMap in supervised mode for baselines and semi-supervised mode for \textit{IsoVec} supervised runs.  
This sometimes underestimates \textit{IsoVec}'s strength when supervised mapping would have performed better.
For unsupervised experiments and baselines, we map in unsupervised mode. Each is run five times and averaged. \textit{IsoVec} and VecMap use one NVIDIA GeForce GTX 1080Ti GPU.

\section{Experiments \& Results}
We pretrain English embeddings to use as reference space $\bf{Y}$. \textit{IsoVec} trains source space $\bf{X}$.
\subsection{Main Experiments}
\label{sec:main}
For baselines, we train source and target spaces separately for each run using our base implementation.  
In experimental conditions, we train the source space with \textit{IsoVec} using each isomorphism loss from Sections \ref{sec:sup_losses} and \ref{sec:unsup_losses}.  
In Table \ref{tab:main}, we see that \textit{IsoVec} consistently outperforms the baseline for bn-en and uk-en.  
For ta-en, it outperforms with \textbf{Proc-L2+Init} and both unsupervised methods.\footnote{Test set coverage *-to-en; bn: 77\%, uk: 76.8\%, ta: 71.4\%} In terms of training efficiency, L2-based methods perform comparably to the baseline (< 10\% time increase) and RSIM-based methods see a slight time increase ($\sim$10-16\% increase over baseline).  EVS-based methods require an expensive eigendecomposition step which causes a $\sim$2.5x time increase over the baseline.

\begin{table}
\centering  \setlength\tabcolsep{3.5pt}\footnotesize
\begin{tabular}{l|ll|ll|ll}
\toprule
                          & \multicolumn{2}{c}{\underline{bn}} & \multicolumn{2}{c}{\underline{uk}} & \multicolumn{2}{c}{\underline{ta}}  \\
\midrule 
\textbf{Supervised} & $\mu$  & $\sigma$  & $\mu$ & $\sigma$ & $\mu$ & $\sigma$ \\
\hspace{1mm}\textit{Baseline}                  & \textit{15.2 }& \textit{(0.8) }& \textit{14.4} & \textit{(0.8) }& \textit{11.6} & \textit{(0.4)}   \\
\hspace{1mm}L2 &   16.3 & (0.4)            &  16.5 & (0.4)  &  11.1 & (0.5)      \\
\hspace{1mm}Proc-L2    &  16.6 & (0.7)  &  16.0 & (0.8)  &  10.7 & (0.3)    \\
\hspace{5mm}+\textit{Init} &  \textbf{16.9} & (0.2)  & \textbf{17.1} & (0.6)   &  \textbf{11.8} & (0.3)   \\
\hspace{1mm}RSIM                      &   16.3 & (0.3)  &  15.9 & (0.4)  &  10.3 & (0.6)   \\
\hspace{5mm}+\textit{Init}           &  16.0 & (0.4)  &  \textbf{17.1} & (0.5)  &  11.0 & (0.4)   \\
\midrule 
\midrule 
\textbf{Unsupervised} & & & &&& \\
\hspace{1mm}\textit{Baseline}   &  13.2 & (0.6)         &  12.6 & (0.5)  &  3.2 & (4.4)   \\
\hspace{1mm}RSIM-U                          &   \textbf{14.2} & (0.7) & \textbf{14.0} & (0.6)   &  \textbf{5.4} & (4.9)   \\
\hspace{1mm}EVS-U                          & 13.4 & (0.7)   &  13.4 & (0.6)  &  5.2 & (4.8)   \\
\bottomrule
\end{tabular}
\caption{Main Experiments. Average P@1 ($\mu$) and standard deviation ($\sigma$) over 5 runs of \textit{IsoVec} with isomorphism losses for bn-en, uk-en, ta-en.}
\label{tab:main}
\end{table}

\subsection{Algorithm, Domain, \& Data Mismatch}
\label{sec:dom-mismatch}
\citet{sogaard-etal-2018-limitations} show that mapping methods fail for embeddings trained with different algorithms,
and that BLI performance deteriorates when source and target domains do not match \citep{marchisio-etal-2020-unsupervised}.  We test \textit{IsoVec} under algorithm and domain mismatch using the best losses from the main experiments: \textbf{Proc-L2+Init} and \textbf{RSIM-U}. We use $\beta$ as-is from the previous section.  

The \textit{IsoVec} base model intends to mirror word2vec closely, but there are likely output differences due to implementation.\footnote{Ours batches on GPU with Adam; word2vec is CPU-only with SGD, no batching.} We map the baseline source embeddings trained in the main experiments to varying en target spaces trained with the official word2vec release, so that algorithms do not match between source and target embedding spaces. We run experiments using the below training data:
\begin{itemize}[]
    \item \textit{Algorithm Mismatch}: 1 million lines of en newscrawl2020 (same as main experiments). Shows effect of algorithm mismatch only.
    \item \textit{+More Target-Side Data}: 93 million lines of en newscrawl2018-20. Shows effect of target trained with ample in-domain data.
    \item \textit{+Domain Mismatch}: 33.8 million lines of en Common Crawl (web-crawl). Shows the effect of different domains in source vs. target.
\end{itemize}

\noindent Table \ref{tab:alg-dom-mismatch-delta} contains baselines for our mismatch experiments and shows the drop in performance compared to Table \ref{tab:main} baselines, where both source and target embedding spaces were trained with the \textit{IsoVec} base model. 
This occurs across languages, moderately for supervised baselines, and severely for unsupervised.
The large performance drop given \textit{more} high-quality data of the same domain in unsupervised mode (\textit{+More Target-Side Data}) is surprising given that this target space is \textit{stronger} than the one from only \textit{Algorithm Mismatch}. Perhaps its geometry has changed so considerably because of its additional data and different algorithm that it is too different from the lower-resource source space to be mapped with unsupervised methods.  
This should be investigated in future work. 

\begin{table*}[htb]
\centering  \setlength\tabcolsep{3pt}\footnotesize 
\begin{tabular}{l|r@{ }lr|r@{ }lr|r@{ }lr||r@{ }lr|r@{ }lr|r@{ }lr}
\toprule
 & \multicolumn{9}{c||}{\underline{Supervised}} & \multicolumn{9}{c}{\underline{Unsupervised}} \\
 & \multicolumn{3}{c}{\underline{bn}} & \multicolumn{3}{c}{\underline{uk}} & \multicolumn{3}{c||}{\underline{ta}} & \multicolumn{3}{c}{\underline{bn}} & \multicolumn{3}{c}{\underline{uk}} & \multicolumn{3}{c}{\underline{ta}} \\
 & \multicolumn{1}{c}{$\mu$} & \multicolumn{1}{c}{$\Delta$} & {$\sigma$} & \multicolumn{1}{c}{$\mu$} & \multicolumn{1}{c}{$\Delta$} & {$\sigma$} & \multicolumn{1}{c}{$\mu$} & \multicolumn{1}{c}{$\Delta$} & {$\sigma$} & \multicolumn{1}{c}{$\mu$} & \multicolumn{1}{c}{$\Delta$} & {$\sigma$} & \multicolumn{1}{c}{$\mu$} & \multicolumn{1}{c}{$\Delta$} & {$\sigma$} & \multicolumn{1}{c}{$\mu$} & \multicolumn{1}{c}{$\Delta$} & {$\sigma$} \\
 \midrule
\textit{Main Baseline} & \textit{15.2} & \multicolumn{1}{c}{\textit{-}} & \textit{0.8} & \textit{14.4} & \multicolumn{1}{c}{\textit{-}} & \textit{0.8} & \textit{11.6} & \multicolumn{1}{c}{\textit{-}} & \textit{0.4} & \textit{13.2} & \multicolumn{1}{c}{\textit{-}} & \textit{0.6} & \textit{12.6} & \multicolumn{1}{c}{\textit{-}} & \textit{0.5} & \textit{3.2} & \multicolumn{1}{c}{\textit{-}} & \textit{4.4} \\
\midrule
Algorithm Mism. & 13.6 & \textcolor{red}{\textit{(-1.6)}} & 0.5 & 11.7 & \textcolor{red}{\textit{(-2.7)}} & 0.6 & 9.4 & \textcolor{red}{\textit{(-2.1)}} & 0.5 & 11.9 & \textcolor{red}{\textit{(-1.3)}} & 0.7 & 4.4 & \textcolor{red}{\textit{(-8.2)}} & 6.1 & 2.6 & \textcolor{red}{\textit{(-0.6)}} & 3.6 \\
\hspace{1mm}+More Trg Data & 16.0 & \textcolor{darkgreen}{\textit{(+0.8)}} & 0.7 & 13.5 & \textcolor{red}{\textit{(-0.9)}} & 0.3 & 11.0 & \textcolor{red}{\textit{(-0.6)}} & 0.6 & 5.9 & \textcolor{red}{\textit{(-7.3)}} & 8.0 & 0.0 & \textcolor{red}{\textit{(-12.6)}} & 0.0 & 0.0 & \textcolor{red}{\textit{(-3.2)}} & 0.0 \\
\hspace{1mm}+Domain Mism. & 10.5 & \textcolor{red}{\textit{(-4.7)}} & 0.6 & 10.0 & \textcolor{red}{\textit{(-4.4)}} & 0.4 & 8.4 & \textcolor{red}{\textit{(-3.2)}} & 0.4 & 0.0 & \textcolor{red}{\textit{(-13.2)}} & 0.0 & 0.0 & \textcolor{red}{\textit{(-12.6)}} & 0.0 & 0.0 & \textcolor{red}{\textit{(-3.2)}} & 0.0 \\
\bottomrule
\end{tabular}
\caption{Effect of algorithm and data mismatch in source vs. target embedding spaces. Average P@1 of 5 runs ($\mu$) with $\Delta$ vs. baseline and std.dev. ($\sigma$). Isomorphism losses are \textit{not} used here. Source-side embeddings are trained with our base implementation, target-side with word2vec (algorithm mismatch). \textit{Main Baseline} is from the main experiments, Table \ref{tab:main}. \textit{+More Target-Side Data} (+More Trg Data) uses nearly 100x more data on the target-side than previous experiments.  \textit{+Domain Mismatch} uses target embeddings trained on $\sim34$M lines of web crawl.}
\label{tab:alg-dom-mismatch-delta}
\end{table*}

We run \textbf{Proc-L2+Init} and \textbf{RSIM-U} in \textit{Algorithm Mismatch}, \textit{+More Target-Side Data}, and \textit{+Domain Mismatch} conditions as described above. Results are in Table \ref{tab:dom-mismatch}. 
In supervised mode, \textit{IsoVec} recovers from algorithm mismatch by 2.7-4.9 points, domain mismatch by 2.5-7.3, and still improves when the target space is trained on $\sim$100x more data. 
Whereas \textit{+Domain Mismatch} and \textit{+More Target-Side Data} baselines fail to extract any correct translation pairs in unsupervised mode, \textbf{RSIM-U} method completely recovers in all conditions: equalling or outperforming the main unsupervised baseline from Table \ref{tab:main} which matched on algorithm, domain, \textit{and} data size.\footnote{Dev set P@1 on \textit{+Domain Mismatch} was near zero despite success on test set.  We note test uses more common words than dev. \citet{czarnowska-etal-2019-dont} find that BLI performance worsens for rarer words, which may have poorly trained embeddings \citep[also,][]{gong2018frage, sogaard-etal-2018-limitations}.}   
\textit{IsoVec} is thus useful for many types of distributional shifts: algorithmic, domain, and amount of data available. 

\begin{table*}[htb]
\centering  \setlength\tabcolsep{3pt}\footnotesize
\begin{tabular}{l||r@{ }l|r@{ }l|r@{ }l||r@{ }l|r@{ }l|r@{ }l||r@{ }l|r@{ }l|r@{ }l}
\toprule
 & \multicolumn{6}{c||}{\underline{Algorithm Mismatch}} & \multicolumn{6}{c||}{\underline{\textit{+ Domain Mismatch}}} & \multicolumn{6}{c}{\underline{\textit{+ More Target-Side Data}}} \\
 & \multicolumn{2}{c}{bn} & \multicolumn{2}{c}{uk} & \multicolumn{2}{c}{ta} & \multicolumn{2}{c}{bn} & \multicolumn{2}{c}{uk} & \multicolumn{2}{c}{ta} & \multicolumn{2}{c}{bn} & \multicolumn{2}{c}{uk} & \multicolumn{2}{c}{ta} \\
 & \multicolumn{1}{c}{$\mu$} & \multicolumn{1}{c}{$\sigma$} & \multicolumn{1}{c}{$\mu$} & \multicolumn{1}{c}{$\sigma$} & \multicolumn{1}{c}{$\mu$} & \multicolumn{1}{c||}{$\sigma$} & \multicolumn{1}{c}{$\mu$} & \multicolumn{1}{c}{$\sigma$} & \multicolumn{1}{c}{$\mu$} & \multicolumn{1}{c}{$\sigma$} &
 \multicolumn{1}{c}{$\mu$} & \multicolumn{1}{c||}{$\sigma$} & \multicolumn{1}{c}{$\mu$} & \multicolumn{1}{c}{$\sigma$} & \multicolumn{1}{c}{$\mu$} & \multicolumn{1}{c}{$\sigma$} &
 \multicolumn{1}{c}{$\mu$} & \multicolumn{1}{c}{$\sigma$} \\
\midrule
\textbf{Supervised} & & & &&& &&&&&&&&\\
\textit{Baseline} & \textit{13.6} & \textit{(0.5}) & \textit{11.7} & \textit{(0.6}) & \textit{9.4} & \textit{(0.5}) & \textit{10.5} & \textit{(0.6}) & \textit{10.0} & \textit{(0.4}) & \textit{8.4} & \textit{(0.4}) & \textit{16.0} & \textit{(0.7}) & \textit{13.5} & \textit{(0.3}) & \textit{11.0} & \textit{(0.6)} \\
\textbf{Proc-L2+I} (Ours) & \textbf{16.3} & (0.4) & \textbf{16.6} & (2.2) & \textbf{12.1} & (0.7) & \textbf{15.5} & (0.7) & \textbf{17.3} & (0.4) & \textbf{10.9} & (0.5) & \textbf{16.2} & (0.4) & \textbf{17.3} & (0.3) & \textbf{11.4} & (0.3) \\
\midrule
\textbf{Unsupervised} & & & &&& &&&&&&&&\\
\textit{Baseline} & \textit{11.9} & \textit{(0.7)} & \textit{4.4} & \textit{(6.1)} & \textit{2.6} & \textit{(3.6)} & \textit{0.0} & \textit{(0.0)} & \textit{0.0} & \textit{(0.0)} & \textit{0.0 }& \textit{(0.0)} & \textit{5.9} & \textit{(8.0)} & \textit{0.0} & \textit{(0.0)} & \textit{0.0} & \textit{(0.0)} \\
\textbf{RSIM-U} (Ours) & \textbf{13.6} & (0.7) & \textbf{13.4} & (1.2) & \textbf{3.6} & (4.8) & \textbf{13.5} & (0.7)  & \textbf{13.1} & (0.8) & \textbf{6.8} & (3.8)  & \textbf{14.4} & (0.7) & \textbf{13.6} & (0.6) & \textbf{3.2} & (4.3) \\
\bottomrule
\end{tabular}
\caption{P@1 of \textit{IsoVec} vs. baseline under algorithm and domain mismatch.  \textit{Baselines} correspond to Table \ref{tab:alg-dom-mismatch-delta} results. Ours, Supervised: \textbf{Proc-L2+Init}. Ours, Unsupervised: \textbf{RSIM-U}.}
\label{tab:dom-mismatch}
\end{table*}

\subsection{Effect on Isomorphism}
\label{sec:effect-on-iso}

Table \ref{tab:final-iso-scores} (left) shows the effect of \textit{IsoVec} on global isomorphism measures. We measure relational similarity, eigenvector similarity, and Gromov-Hausdorff distance of trained embedding spaces (before mapping) for all main experiments of Section \ref{sec:main} using scripts from \citet{vulic-etal-2020-good}\footnote{\url{https://github.com/cambridgeltl/iso-study/tree/master/scripts}}. We average over experiments.  To avoid confusion with the \textit{IsoVec} loss functions, we call the metrics ``RelSim'', ``EigSim'', and ``GH''. The script calculates EigSim ($k=2$) over the first 10,000 embeddings in each space and GH over the first 5000. RelSim is calculated over the first 1000 seed translation pairs.

All supervised methods improved RelSim ($\uparrow$ better). Perhaps surprisingly, initializing the source space with target embeddings (\textbf{+Init}) worsens isomorphism.  \textbf{RSIM} is best, directly optimizing for this metric in a supervised manner.  RelSim stayed roughly consistent in unsupervised experiments.   

All uk-en and ta-en experiments improve GH \cite[$\downarrow$ better]{patra-etal-2019-bilingual}. GH worsened for bn-en despite improved BLI (Table \ref{tab:main}).  
EigSim ( $\downarrow$ better) improves across all experiments except uk-en supervised methods, despite improved BLI (notably, initial EigSim for uk was low).  \textbf{EVS-U} strongly improves EigSim, optimizing it directly. Table \ref{tab:final-iso-scores} (left) measures the unperturbed geometry of spaces after training and shows that \textit{IsoVec} improves isomorphism in a majority of settings. The same calculation over embeddings \textit{after} mapping with semi-supervised VecMap is in Table \ref{app:final-iso-scores-mapped}. 

It is interesting that baseline experiments performed better when mapped in supervised mode while spaces trained with \textit{IsoVec} tended to map better in semi-supervised mode (as mentioned in Section \ref{sec:beta}). This may further indicate that the \textit{IsoVec} spaces have become more geometrically similar. 

\begin{table*}[!htb]
    \setlength\tabcolsep{3.2pt}\footnotesize
    \begin{minipage}{.56\linewidth}
        \begin{tabular}{@{}l|ccc|ccc|ccc}
            \toprule
            & \multicolumn{3}{c}{\underline{Relational Sim. $\uparrow$}} & \multicolumn{3}{c}{\underline{GH Distance $\downarrow$}} & \multicolumn{3}{c}{\underline{Eigenvector Sim. $\downarrow$}} \\
            & bn & uk & ta & bn & uk & ta & bn & uk & ta \\
            \midrule
            \hspace{1mm}\textit{Baseline} & \textit{0.32} & \textit{0.26} & \textit{0.27} & \textit{\textbf{0.34}} & \textit{1.17} & \textit{0.39} & \textit{62.7} & \textit{49.9} & \textit{74.7} \\
            \hspace{1mm}\textbf{L2} & 0.36 & 0.34 & 0.33 & 0.43 & 0.76 & \textbf{0.19} & 52.0 & 68.5 & 47.3 \\
            \hspace{1mm}\textbf{Proc-L2} & 0.43 & 0.42 & 0.39 & 0.42 & 0.51 & 0.23 & 46.6 & 72.8 & \textbf{37.7} \\
            \hspace{5mm}+\textit{\textbf{Init}} & 0.39 & 0.38 & 0.35 & 0.43 & \textbf{0.46} & 0.22 & 52.1 & 77.2 & 41.0 \\
            \hspace{1mm}\textbf{RSIM}  & \textbf{0.54} & \textbf{0.53} & \textbf{0.47} & 0.37 & 0.56 & 0.20 & 47.2 & 60.4 & 39.8 \\
            \hspace{5mm}+\textit{\textbf{Init}}  & 0.37 & 0.35 & 0.33 & 0.40 & 0.52 & 0.20 & 48.6 & 65.6 & 44.8 \\
            \hspace{1mm}\textbf{RSIM-U} & 0.30 & 0.25 & 0.26 & 0.42 & 0.54 & 0.32 & 56.3 & \textbf{36.2} & 60.1 \\
            \hspace{1mm}\textbf{EVS-U}  & 0.30 & 0.24 & 0.26 & 0.68 & 0.53 & 0.30 & \textbf{38.7} & 38.3 & 39.5 \\
            \bottomrule
        \end{tabular}
    \end{minipage}%
    \begin{minipage}{.44\linewidth}
        \;\,
        \begin{tabular}{l|c|c|rrr}
            \toprule
             & & All & bn-en & uk-en & ta-en \\
             \midrule
            P@1 \hspace{1mm} vs. RelSim & (+) & 0.17 & \textcolor{gray}{\textit{-0.03}} & \textcolor{red}{-0.54} & \textcolor{red}{-0.54} \\
            \hspace{8mm} vs. GH & (-) & \textcolor{red}{0.79} & \textcolor{red}{0.28} & -0.23 & \textcolor{red}{0.27} \\
            \hspace{8mm} vs. EigSim & (-) & \textcolor{red}{0.57} & \textcolor{red}{0.17} & \textcolor{gray}{\textit{-0.05}} & \textcolor{gray}{\textit{0.05}} \\
            \midrule
            RelSim \hspace{2mm} vs. GH & (-) & \textcolor{red}{0.09} & -0.45 & -0.24 & \textcolor{gray}{\textit{-0.03}} \\
            \hspace{8mm} vs. EigSim & (-) & \textcolor{gray}{\textit{-0.05}} & -0.25 & -0.28 & -0.27 \\
            GH  \hspace{2.5mm} vs. EigSim & (+) & 0.65 & \textcolor{red}{-0.16} & \textcolor{red}{-0.17} & \textcolor{red}{-0.29} \\
            \bottomrule
            \end{tabular}
    \end{minipage} 
    \caption{\textbf{Left:} Isomorphism scores of source vs. target embedding space after training, per language, per training method, averaged over 5 runs.  Best is \textbf{bold}. \textit{IsoVec} output embeddings vs. base English space. 
    \textbf{Right:} Pearson's correlation: BLI performance (P@1) of supervised \textit{IsoVec} experiments vs. isomorphism score 
    and between metrics. (+/-): expected direction of correlation: \textcolor{red}{opposite} or \textit{\textcolor{gray}{weak}} (magnitude <= 0.05).
    }
\label{tab:final-iso-scores}
\label{tab:corrs}
\end{table*}

\section{Discussion}
\subsection{The Promise of Geometric Losses}
We have seen that \textit{IsoVec} improves relative isomorphism and downstream BLI from word embedding spaces.
The success of unsupervised methods is particularly encouraging for the use of global isomorphism measures to improve embedding spaces. 
Notably, we use only the first 2000 words per space to calculate unsupervised \textit{IsoVec} losses---i.e., we coax these frequent words to have similar representations, regardless of identity. 
While there are likely some true translation pairs in the mix, there are almost certainly words this subset of $\bf{X}$ whose translation is not in the first 2000 words of $\bf{Y}$ (and vice-versa)---particularly when source and target corpora are from different domains.  
Regardless, \textit{IsoVec} unsupervised methods work. 

\subsection{Need for a Sensitive Isomorphism Metric}
Previous authors found that EigSim and GH correlate well with BLI performance \cite{sogaard-etal-2018-limitations, patra-etal-2019-bilingual}, 
however our results reveal a nuanced story. 
In Table \ref{tab:corrs} (right), we correlate the EigSim, RelSim, and GH with BLI P@1 performance over all runs of the main supervised \textit{IsoVec} experiments (\textbf{L2}, \textbf{Proc-L2}, \textbf{Proc-L2+Init}, \textbf{RSIM}, \textbf{RSIM+Init}; 25 data points per calculation).

P@1 should correlate positively with RelSim ($\uparrow$ better for both) and negatively vs. GH/EigSim ($\downarrow$ better for GH/EigSim).  
RelSim should correlate negatively with GH/EigSim, and GH positively with EigSim.  
In Table \ref{tab:corrs} (right) within language, however, only P@1 vs. GH on uk-en aligns with intuition.  
Many correlations are weak (gray, magnitude <= 0.05) or \textit{opposite} of expected;  For instance, P@1 should increase with RelSim, but we see the opposite within language pair.  
Over languages combined, the relationship is weakly positive. Figure \ref{fig:rsim-v-p1} shows how this is possible.  

Samples for Pearson's correlation should be drawn from the same population, and in Table \ref{tab:corrs} (right) we assume that is our \textit{IsoVec} embedding spaces. 
Perhaps the assumption is unfair: different \textit{IsoVec} losses might induce different monolingual spaces where specific metrics are indeed predictive of downstream BLI performance, but this may not be visible in the aggregate.
An ideal metric, however, would predict downstream BLI performance regardless of how monolingual spaces were trained; such that we might assess the potential of spaces to align well without having to map them and measure their performance with development or test dictionaries.
In that light, the discrepancies in Table \ref{tab:corrs} (right) highlight the need for a more sensitive metric that works within language and with small differences in BLI performance.\footnote{For instance, the five runs of \textbf{L2} ranged in P@1 from 16.05-16.91, and in RelSim from 0.3606-0.3618.
Maximum and minimum BLI scores differed by only 0.0004 RSIM.
}

We should thus be cautious drawing between- vs. within-language conclusions about isomorphism metrics and downstream BLI.  
When isomorphism metrics differ considerably, perhaps BLI performance also differs similarly, as seen in previous work; however if isomorphism scores are poor or too similar, the metrics may not be sensitive enough to be predictive. Future work should investigate these hypotheses and develop isomorphism metrics that are more sensitive. The spectral measures of \citet{dubossarsky-etal-2020-secret} might be examined in these lower-resource contexts, as the authors claim to correlate better with downstream BLI. 
All-in-all, though, our main results show that coaxing towards improved isomorphism as measured by the three popular metrics can improve BLI performance even if the scores are not strongly predictive of raw P@1.  

\section{Conclusion \& Future Work}
We present \textit{IsoVec}, a new method for training word embeddings which directly injects global measures of embedding space isomorphism into the Skip-gram loss function. 
Our three supervised and two unsupervised isomorphism loss functions successfully improve the mappability of monolingual word embedding spaces, leading to improved ability to induce bilingual lexicons.  
\textit{IsoVec} also shows promise under algorithm mismatch, domain mismatch, and data size mismatch between source and target training corpora.  
Future work could extend our work to even greater algorithmic mismatches, and in massively multilingual contextualized models. 
We release \textit{IsoVec} at \url{https://github.com/kellymarchisio/isovec}. 

\section*{Limitations}
As with most methods based on static word embeddings, our work is limited by polysemy. 
By using word2vec as a basis, we inherit many of its limitations, many of which are addressed in recent contextualized representation learning work. 
Future work might apply our methods to contextualized models. 
We also experiment with only English as a target language, limiting our method's universal applicability. 
Future work could extend our results to non-English pairs, and also evaluate monolingually if languages will be used separately as recommended by \citet{luong-etal-2015-bilingual}. 

\section*{Acknowledgements}
This material is based upon work supported by the United States Air Force under Contract No. FA8750‐19‐C‐0098.  Any opinions, findings, and conclusions or recommendations expressed in this material are those of the author(s) and do not necessarily reflect the views of the United States Air Force and DARPA.

\bibliography{anthology,custom}

\begin{thebibliography}{59}
\expandafter\ifx\csname natexlab\endcsname\relax\def\natexlab#1{#1}\fi

\bibitem[{Alvarez-Melis and
  Jaakkola(2018)}]{alvarez-melis-jaakkola-2018-gromov}
David Alvarez-Melis and Tommi Jaakkola. 2018.
\newblock \href {https://doi.org/10.18653/v1/D18-1214} {{G}romov-{W}asserstein
  alignment of word embedding spaces}.
\newblock In \emph{Proceedings of the 2018 Conference on Empirical Methods in
  Natural Language Processing}, pages 1881--1890, Brussels, Belgium.
  Association for Computational Linguistics.

\bibitem[{Artetxe et~al.(2016)Artetxe, Labaka, and
  Agirre}]{artetxe-etal-2016-learning}
Mikel Artetxe, Gorka Labaka, and Eneko Agirre. 2016.
\newblock \href {https://doi.org/10.18653/v1/D16-1250} {Learning principled
  bilingual mappings of word embeddings while preserving monolingual
  invariance}.
\newblock In \emph{Proceedings of the 2016 Conference on Empirical Methods in
  Natural Language Processing}, pages 2289--2294, Austin, Texas. Association
  for Computational Linguistics.

\bibitem[{Artetxe et~al.(2017)Artetxe, Labaka, and
  Agirre}]{artetxe-etal-2017-learning}
Mikel Artetxe, Gorka Labaka, and Eneko Agirre. 2017.
\newblock \href {https://doi.org/10.18653/v1/P17-1042} {Learning bilingual word
  embeddings with (almost) no bilingual data}.
\newblock In \emph{Proceedings of the 55th Annual Meeting of the Association
  for Computational Linguistics (Volume 1: Long Papers)}, pages 451--462,
  Vancouver, Canada. Association for Computational Linguistics.

\bibitem[{Artetxe et~al.(2018{\natexlab{a}})Artetxe, Labaka, and
  Agirre}]{artetxe2018generalizing}
Mikel Artetxe, Gorka Labaka, and Eneko Agirre. 2018{\natexlab{a}}.
\newblock Generalizing and improving bilingual word embedding mappings with a
  multi-step framework of linear transformations.
\newblock In \emph{Proceedings of the AAAI Conference on Artificial
  Intelligence}, volume~32.

\bibitem[{Artetxe et~al.(2018{\natexlab{b}})Artetxe, Labaka, and
  Agirre}]{artetxe-etal-2018-robust}
Mikel Artetxe, Gorka Labaka, and Eneko Agirre. 2018{\natexlab{b}}.
\newblock \href {https://doi.org/10.18653/v1/P18-1073} {A robust self-learning
  method for fully unsupervised cross-lingual mappings of word embeddings}.
\newblock In \emph{Proceedings of the 56th Annual Meeting of the Association
  for Computational Linguistics (Volume 1: Long Papers)}, pages 789--798,
  Melbourne, Australia. Association for Computational Linguistics.

\bibitem[{Barrault et~al.(2020)Barrault, Biesialska, Bojar, Costa-juss{\`a},
  Federmann, Graham, Grundkiewicz, Haddow, Huck, Joanis, Kocmi, Koehn, Lo,
  Ljube{\v{s}}i{\'c}, Monz, Morishita, Nagata, Nakazawa, Pal, Post, and
  Zampieri}]{barrault-etal-2020-findings}
Lo{\"\i}c Barrault, Magdalena Biesialska, Ond{\v{r}}ej Bojar, Marta~R.
  Costa-juss{\`a}, Christian Federmann, Yvette Graham, Roman Grundkiewicz,
  Barry Haddow, Matthias Huck, Eric Joanis, Tom Kocmi, Philipp Koehn, Chi-kiu
  Lo, Nikola Ljube{\v{s}}i{\'c}, Christof Monz, Makoto Morishita, Masaaki
  Nagata, Toshiaki Nakazawa, Santanu Pal, Matt Post, and Marcos Zampieri. 2020.
\newblock \href {https://aclanthology.org/2020.wmt-1.1} {Findings of the 2020
  conference on machine translation ({WMT}20)}.
\newblock In \emph{Proceedings of the Fifth Conference on Machine Translation},
  pages 1--55, Online. Association for Computational Linguistics.

\bibitem[{Cao et~al.(2020)Cao, Kitaev, and Klein}]{Cao2020Multilingual}
Steven Cao, Nikita Kitaev, and Dan Klein. 2020.
\newblock \href {https://openreview.net/forum?id=r1xCMyBtPS} {Multilingual
  alignment of contextual word representations}.
\newblock In \emph{International Conference on Learning Representations}.

\bibitem[{Chazal et~al.(2009)Chazal, Cohen-Steiner, Guibas, Mémoli, and
  Oudot}]{chazal-2009-bottleneck}
Frédéric Chazal, David Cohen-Steiner, Leonidas~J. Guibas, Facundo Mémoli,
  and Steve~Y. Oudot. 2009.
\newblock \href
  {https://doi.org/https://doi.org/10.1111/j.1467-8659.2009.01516.x}
  {Gromov-hausdorff stable signatures for shapes using persistence}.
\newblock \emph{Computer Graphics Forum}, 28(5):1393--1403.

\bibitem[{Conneau et~al.(2020)Conneau, Khandelwal, Goyal, Chaudhary, Wenzek,
  Guzm{\'a}n, Grave, Ott, Zettlemoyer, and
  Stoyanov}]{conneau-etal-2020-unsupervised}
Alexis Conneau, Kartikay Khandelwal, Naman Goyal, Vishrav Chaudhary, Guillaume
  Wenzek, Francisco Guzm{\'a}n, Edouard Grave, Myle Ott, Luke Zettlemoyer, and
  Veselin Stoyanov. 2020.
\newblock \href {https://doi.org/10.18653/v1/2020.acl-main.747} {Unsupervised
  cross-lingual representation learning at scale}.
\newblock In \emph{Proceedings of the 58th Annual Meeting of the Association
  for Computational Linguistics}, pages 8440--8451, Online. Association for
  Computational Linguistics.

\bibitem[{Czarnowska et~al.(2019)Czarnowska, Ruder, Grave, Cotterell, and
  Copestake}]{czarnowska-etal-2019-dont}
Paula Czarnowska, Sebastian Ruder, Edouard Grave, Ryan Cotterell, and Ann
  Copestake. 2019.
\newblock \href {https://doi.org/10.18653/v1/D19-1090} {Don{'}t forget the long
  tail! a comprehensive analysis of morphological generalization in bilingual
  lexicon induction}.
\newblock In \emph{Proceedings of the 2019 Conference on Empirical Methods in
  Natural Language Processing and the 9th International Joint Conference on
  Natural Language Processing (EMNLP-IJCNLP)}, pages 974--983, Hong Kong,
  China. Association for Computational Linguistics.

\bibitem[{Devlin et~al.(2019)Devlin, Chang, Lee, and
  Toutanova}]{devlin-etal-2019-bert}
Jacob Devlin, Ming-Wei Chang, Kenton Lee, and Kristina Toutanova. 2019.
\newblock \href {https://doi.org/10.18653/v1/N19-1423} {{BERT}: Pre-training of
  deep bidirectional transformers for language understanding}.
\newblock In \emph{Proceedings of the 2019 Conference of the North {A}merican
  Chapter of the Association for Computational Linguistics: Human Language
  Technologies, Volume 1 (Long and Short Papers)}, pages 4171--4186,
  Minneapolis, Minnesota. Association for Computational Linguistics.

\bibitem[{Doval et~al.(2018)Doval, Camacho-Collados, Espinosa-Anke, and
  Schockaert}]{doval-etal-2018-improving}
Yerai Doval, Jose Camacho-Collados, Luis Espinosa-Anke, and Steven Schockaert.
  2018.
\newblock \href {https://doi.org/10.18653/v1/D18-1027} {Improving cross-lingual
  word embeddings by meeting in the middle}.
\newblock In \emph{Proceedings of the 2018 Conference on Empirical Methods in
  Natural Language Processing}, pages 294--304, Brussels, Belgium. Association
  for Computational Linguistics.

\bibitem[{Dubossarsky et~al.(2020)Dubossarsky, Vuli{\'c}, Reichart, and
  Korhonen}]{dubossarsky-etal-2020-secret}
Haim Dubossarsky, Ivan Vuli{\'c}, Roi Reichart, and Anna Korhonen. 2020.
\newblock \href {https://doi.org/10.18653/v1/2020.emnlp-main.186} {The secret
  is in the spectra: Predicting cross-lingual task performance with spectral
  similarity measures}.
\newblock In \emph{Proceedings of the 2020 Conference on Empirical Methods in
  Natural Language Processing (EMNLP)}, pages 2377--2390, Online. Association
  for Computational Linguistics.

\bibitem[{Eder et~al.(2021)Eder, Hangya, and Fraser}]{eder-etal-2021-anchor}
Tobias Eder, Viktor Hangya, and Alexander Fraser. 2021.
\newblock \href {https://doi.org/10.18653/v1/2021.acl-short.30} {Anchor-based
  bilingual word embeddings for low-resource languages}.
\newblock In \emph{Proceedings of the 59th Annual Meeting of the Association
  for Computational Linguistics and the 11th International Joint Conference on
  Natural Language Processing (Volume 2: Short Papers)}, pages 227--232,
  Online. Association for Computational Linguistics.

\bibitem[{Ethayarajh(2019)}]{ethayarajh-2019-contextual}
Kawin Ethayarajh. 2019.
\newblock \href {https://doi.org/10.18653/v1/D19-1006} {How contextual are
  contextualized word representations? {C}omparing the geometry of {BERT},
  {ELM}o, and {GPT}-2 embeddings}.
\newblock In \emph{Proceedings of the 2019 Conference on Empirical Methods in
  Natural Language Processing and the 9th International Joint Conference on
  Natural Language Processing (EMNLP-IJCNLP)}, pages 55--65, Hong Kong, China.
  Association for Computational Linguistics.

\bibitem[{Ethayarajh and Jurafsky(2021)}]{ethayarajh-jurafsky-2021-attention}
Kawin Ethayarajh and Dan Jurafsky. 2021.
\newblock \href {https://doi.org/10.18653/v1/2021.acl-short.8} {Attention flows
  are shapley value explanations}.
\newblock In \emph{Proceedings of the 59th Annual Meeting of the Association
  for Computational Linguistics and the 11th International Joint Conference on
  Natural Language Processing (Volume 2: Short Papers)}, pages 49--54, Online.
  Association for Computational Linguistics.

\bibitem[{Faruqui et~al.(2015)Faruqui, Dodge, Jauhar, Dyer, Hovy, and
  Smith}]{faruqui-etal-2015-retrofitting}
Manaal Faruqui, Jesse Dodge, Sujay~Kumar Jauhar, Chris Dyer, Eduard Hovy, and
  Noah~A. Smith. 2015.
\newblock \href {https://doi.org/10.3115/v1/N15-1184} {Retrofitting word
  vectors to semantic lexicons}.
\newblock In \emph{Proceedings of the 2015 Conference of the North {A}merican
  Chapter of the Association for Computational Linguistics: Human Language
  Technologies}, pages 1606--1615, Denver, Colorado. Association for
  Computational Linguistics.

\bibitem[{Glava{\v{s}} et~al.(2019)Glava{\v{s}}, Litschko, Ruder, and
  Vuli{\'c}}]{glavas-etal-2019-properly}
Goran Glava{\v{s}}, Robert Litschko, Sebastian Ruder, and Ivan Vuli{\'c}. 2019.
\newblock \href {https://doi.org/10.18653/v1/P19-1070} {How to (properly)
  evaluate cross-lingual word embeddings: On strong baselines, comparative
  analyses, and some misconceptions}.
\newblock In \emph{Proceedings of the 57th Annual Meeting of the Association
  for Computational Linguistics}, pages 710--721, Florence, Italy. Association
  for Computational Linguistics.

\bibitem[{Glava{\v{s}} and Vuli{\'c}(2020)}]{glavas-vulic-2020-non}
Goran Glava{\v{s}} and Ivan Vuli{\'c}. 2020.
\newblock \href {https://doi.org/10.18653/v1/2020.acl-main.675} {Non-linear
  instance-based cross-lingual mapping for non-isomorphic embedding spaces}.
\newblock In \emph{Proceedings of the 58th Annual Meeting of the Association
  for Computational Linguistics}, pages 7548--7555, Online. Association for
  Computational Linguistics.

\bibitem[{Gong et~al.(2018)Gong, He, Tan, Qin, Wang, and Liu}]{gong2018frage}
Chengyue Gong, Di~He, Xu~Tan, Tao Qin, Liwei Wang, and Tie-Yan Liu. 2018.
\newblock Frage: Frequency-agnostic word representation.
\newblock \emph{Advances in neural information processing systems}, 31.

\bibitem[{Gouws et~al.(2015)Gouws, Bengio, and Corrado}]{gouws-2015}
Stephan Gouws, Yoshua Bengio, and Greg Corrado. 2015.
\newblock Bilbowa: Fast bilingual distributed representations without word
  alignments.
\newblock In \emph{Proceedings of the 32nd International Conference on
  International Conference on Machine Learning - Volume 37}, ICML'15, page
  748–756. JMLR.org.

\bibitem[{Jawanpuria et~al.(2019)Jawanpuria, Balgovind, Kunchukuttan, and
  Mishra}]{jawanpuria-etal-2019-learning}
Pratik Jawanpuria, Arjun Balgovind, Anoop Kunchukuttan, and Bamdev Mishra.
  2019.
\newblock \href {https://doi.org/10.1162/tacl_a_00257} {Learning multilingual
  word embeddings in latent metric space: A geometric approach}.
\newblock \emph{Transactions of the Association for Computational Linguistics},
  7:107--120.

\bibitem[{Joulin et~al.(2018)Joulin, Bojanowski, Mikolov, J{\'e}gou, and
  Grave}]{joulin-etal-2018-loss}
Armand Joulin, Piotr Bojanowski, Tomas Mikolov, Herv{\'e} J{\'e}gou, and
  Edouard Grave. 2018.
\newblock \href {https://doi.org/10.18653/v1/D18-1330} {Loss in translation:
  Learning bilingual word mapping with a retrieval criterion}.
\newblock In \emph{Proceedings of the 2018 Conference on Empirical Methods in
  Natural Language Processing}, pages 2979--2984, Brussels, Belgium.
  Association for Computational Linguistics.

\bibitem[{Kessy et~al.(2018)Kessy, Lewin, and Strimmer}]{kessy2018optimal}
Agnan Kessy, Alex Lewin, and Korbinian Strimmer. 2018.
\newblock Optimal whitening and decorrelation.
\newblock \emph{The American Statistician}, 72(4):309--314.

\bibitem[{Kim et~al.(2020)Kim, Gra{\c{c}}a, and Ney}]{kim2020and}
Yunsu Kim, Miguel Gra{\c{c}}a, and Hermann Ney. 2020.
\newblock When and why is unsupervised neural machine translation useless?
\newblock In \emph{Proceedings of the 22nd Annual Conference of the European
  Association for Machine Translation}, pages 35--44.

\bibitem[{Kulshreshtha et~al.(2020)Kulshreshtha, Redondo~Garcia, and
  Chang}]{kulshreshtha-etal-2020-cross}
Saurabh Kulshreshtha, Jose~Luis Redondo~Garcia, and Ching-Yun Chang. 2020.
\newblock \href {https://doi.org/10.18653/v1/2020.findings-emnlp.83}
  {Cross-lingual alignment methods for multilingual {BERT}: A comparative
  study}.
\newblock In \emph{Findings of the Association for Computational Linguistics:
  EMNLP 2020}, pages 933--942, Online. Association for Computational
  Linguistics.

\bibitem[{Lample et~al.(2018)Lample, Conneau, Ranzato, Denoyer, and
  J{\'e}gou}]{lample2018word}
Guillaume Lample, Alexis Conneau, Marc'Aurelio Ranzato, Ludovic Denoyer, and
  Herv{\'e} J{\'e}gou. 2018.
\newblock Word translation without parallel data.
\newblock In \emph{International Conference on Learning Representations}.

\bibitem[{Li et~al.(2020)Li, Zhou, He, Wang, Yang, and
  Li}]{li-etal-2020-sentence}
Bohan Li, Hao Zhou, Junxian He, Mingxuan Wang, Yiming Yang, and Lei Li. 2020.
\newblock \href {https://doi.org/10.18653/v1/2020.emnlp-main.733} {On the
  sentence embeddings from pre-trained language models}.
\newblock In \emph{Proceedings of the 2020 Conference on Empirical Methods in
  Natural Language Processing (EMNLP)}, pages 9119--9130, Online. Association
  for Computational Linguistics.

\bibitem[{Li et~al.(2022)Li, Liu, Collier, Korhonen, and
  Vuli{\'c}}]{li-etal-2022-improving}
Yaoyiran Li, Fangyu Liu, Nigel Collier, Anna Korhonen, and Ivan Vuli{\'c}.
  2022.
\newblock \href {https://aclanthology.org/2022.acl-long.299} {Improving word
  translation via two-stage contrastive learning}.
\newblock In \emph{Proceedings of the 60th Annual Meeting of the Association
  for Computational Linguistics (Volume 1: Long Papers)}, pages 4353--4374,
  Dublin, Ireland. Association for Computational Linguistics.

\bibitem[{Liu et~al.(2019)Liu, McCarthy, Vuli{\'c}, and
  Korhonen}]{liu-etal-2019-investigating}
Qianchu Liu, Diana McCarthy, Ivan Vuli{\'c}, and Anna Korhonen. 2019.
\newblock \href {https://doi.org/10.18653/v1/K19-1004} {Investigating
  cross-lingual alignment methods for contextualized embeddings with
  token-level evaluation}.
\newblock In \emph{Proceedings of the 23rd Conference on Computational Natural
  Language Learning (CoNLL)}, pages 33--43, Hong Kong, China. Association for
  Computational Linguistics.

\bibitem[{Luong et~al.(2015)Luong, Pham, and
  Manning}]{luong-etal-2015-bilingual}
Thang Luong, Hieu Pham, and Christopher~D. Manning. 2015.
\newblock \href {https://doi.org/10.3115/v1/W15-1521} {Bilingual word
  representations with monolingual quality in mind}.
\newblock In \emph{Proceedings of the 1st Workshop on Vector Space Modeling for
  Natural Language Processing}, pages 151--159, Denver, Colorado. Association
  for Computational Linguistics.

\bibitem[{Marchisio et~al.(2020)Marchisio, Duh, and
  Koehn}]{marchisio-etal-2020-unsupervised}
Kelly Marchisio, Kevin Duh, and Philipp Koehn. 2020.
\newblock \href {https://aclanthology.org/2020.wmt-1.68} {When does
  unsupervised machine translation work?}
\newblock In \emph{Proceedings of the Fifth Conference on Machine Translation},
  pages 571--583, Online. Association for Computational Linguistics.

\bibitem[{Marie and Fujita(2020)}]{marie-unsup-mt-2020}
Benjamin Marie and Atsushi Fujita. 2020.
\newblock \href {https://doi.org/10.1145/3389790} {Iterative training of
  unsupervised neural and statistical machine translation systems}.
\newblock \emph{ACM Trans. Asian Low-Resour. Lang. Inf. Process.}, 19(5).

\bibitem[{Miceli~Barone(2016)}]{miceli-barone-2016-towards}
Antonio~Valerio Miceli~Barone. 2016.
\newblock \href {https://doi.org/10.18653/v1/W16-1614} {Towards cross-lingual
  distributed representations without parallel text trained with adversarial
  autoencoders}.
\newblock In \emph{Proceedings of the 1st Workshop on Representation Learning
  for {NLP}}, pages 121--126, Berlin, Germany. Association for Computational
  Linguistics.

\bibitem[{Mikolov et~al.(2013{\natexlab{a}})Mikolov, Chen, Corrado, and
  Dean}]{mikolov2013efficient}
Tomas Mikolov, Kai Chen, Greg Corrado, and Jeffrey Dean. 2013{\natexlab{a}}.
\newblock Efficient estimation of word representations in vector space.
\newblock \emph{arXiv preprint arXiv:1301.3781}.

\bibitem[{Mikolov et~al.(2013{\natexlab{b}})Mikolov, Sutskever, Chen, Corrado,
  and Dean}]{mikolov2013}
Tomas Mikolov, Ilya Sutskever, Kai Chen, Greg~S Corrado, and Jeff Dean.
  2013{\natexlab{b}}.
\newblock Distributed representations of words and phrases and their
  compositionality.
\newblock In \emph{Advances in neural information processing systems}, pages
  3111--3119.

\bibitem[{Mimno and Thompson(2017)}]{mimno-thompson-2017-strange}
David Mimno and Laure Thompson. 2017.
\newblock \href {https://doi.org/10.18653/v1/D17-1308} {The strange geometry of
  skip-gram with negative sampling}.
\newblock In \emph{Proceedings of the 2017 Conference on Empirical Methods in
  Natural Language Processing}, pages 2873--2878, Copenhagen, Denmark.
  Association for Computational Linguistics.

\bibitem[{Mohiuddin et~al.(2020)Mohiuddin, Bari, and
  Joty}]{mohiuddin-etal-2020-lnmap}
Tasnim Mohiuddin, M~Saiful Bari, and Shafiq Joty. 2020.
\newblock \href {https://doi.org/10.18653/v1/2020.emnlp-main.215} {{LNM}ap:
  Departures from isomorphic assumption in bilingual lexicon induction through
  non-linear mapping in latent space}.
\newblock In \emph{Proceedings of the 2020 Conference on Empirical Methods in
  Natural Language Processing (EMNLP)}, pages 2712--2723, Online. Association
  for Computational Linguistics.

\bibitem[{Mu and Viswanath(2018)}]{mu2018allbutthetop}
Jiaqi Mu and Pramod Viswanath. 2018.
\newblock \href {https://openreview.net/forum?id=HkuGJ3kCb} {All-but-the-top:
  Simple and effective postprocessing for word representations}.
\newblock In \emph{International Conference on Learning Representations}.

\bibitem[{Nakashole and Flauger(2018)}]{nakashole-flauger-2018-characterizing}
Ndapa Nakashole and Raphael Flauger. 2018.
\newblock \href {https://doi.org/10.18653/v1/P18-2036} {Characterizing
  departures from linearity in word translation}.
\newblock In \emph{Proceedings of the 56th Annual Meeting of the Association
  for Computational Linguistics (Volume 2: Short Papers)}, pages 221--227,
  Melbourne, Australia. Association for Computational Linguistics.

\bibitem[{Ormazabal et~al.(2019)Ormazabal, Artetxe, Labaka, Soroa, and
  Agirre}]{ormazabal-etal-2019-analyzing}
Aitor Ormazabal, Mikel Artetxe, Gorka Labaka, Aitor Soroa, and Eneko Agirre.
  2019.
\newblock \href {https://doi.org/10.18653/v1/P19-1492} {Analyzing the
  limitations of cross-lingual word embedding mappings}.
\newblock In \emph{Proceedings of the 57th Annual Meeting of the Association
  for Computational Linguistics}, pages 4990--4995, Florence, Italy.
  Association for Computational Linguistics.

\bibitem[{Ormazabal et~al.(2021)Ormazabal, Artetxe, Soroa, Labaka, and
  Agirre}]{ormazabal-etal-2021-beyond}
Aitor Ormazabal, Mikel Artetxe, Aitor Soroa, Gorka Labaka, and Eneko Agirre.
  2021.
\newblock \href {https://doi.org/10.18653/v1/2021.acl-long.506} {Beyond offline
  mapping: Learning cross-lingual word embeddings through context anchoring}.
\newblock In \emph{Proceedings of the 59th Annual Meeting of the Association
  for Computational Linguistics and the 11th International Joint Conference on
  Natural Language Processing (Volume 1: Long Papers)}, pages 6479--6489,
  Online. Association for Computational Linguistics.

\bibitem[{Patra et~al.(2019)Patra, Moniz, Garg, Gormley, and
  Neubig}]{patra-etal-2019-bilingual}
Barun Patra, Joel Ruben~Antony Moniz, Sarthak Garg, Matthew~R. Gormley, and
  Graham Neubig. 2019.
\newblock \href {https://doi.org/10.18653/v1/P19-1018} {Bilingual lexicon
  induction with semi-supervision in non-isometric embedding spaces}.
\newblock In \emph{Proceedings of the 57th Annual Meeting of the Association
  for Computational Linguistics}, pages 184--193, Florence, Italy. Association
  for Computational Linguistics.

\bibitem[{Peng et~al.(2021)Peng, Lin, and Stevenson}]{peng-etal-2021-cross}
Xutan Peng, Chenghua Lin, and Mark Stevenson. 2021.
\newblock \href {https://doi.org/10.18653/v1/2021.naacl-main.214}
  {Cross-lingual word embedding refinement by $\ell_{1}$ norm optimisation}.
\newblock In \emph{Proceedings of the 2021 Conference of the North American
  Chapter of the Association for Computational Linguistics: Human Language
  Technologies}, pages 2690--2701, Online. Association for Computational
  Linguistics.

\bibitem[{Rajaee and Pilehvar(2022)}]{rajaee-pilehvar-2022-isotropy}
Sara Rajaee and Mohammad~Taher Pilehvar. 2022.
\newblock \href {https://aclanthology.org/2022.findings-acl.103} {An isotropy
  analysis in the multilingual {BERT} embedding space}.
\newblock In \emph{Findings of the Association for Computational Linguistics:
  ACL 2022}, pages 1309--1316, Dublin, Ireland. Association for Computational
  Linguistics.

\bibitem[{Ruder et~al.(2019)Ruder, Vuli{\'c}, and S{\o}gaard}]{ruder2019survey}
Sebastian Ruder, Ivan Vuli{\'c}, and Anders S{\o}gaard. 2019.
\newblock \href {https://doi.org/10.1613/jair.1.11640} {A survey of
  cross-lingual word embedding models}.
\newblock \emph{Journal of Artificial Intelligence Research}, 65:569--631.

\bibitem[{Rudman et~al.(2022)Rudman, Gillman, Rayne, and
  Eickhoff}]{rudman-etal-2022-isoscore}
William Rudman, Nate Gillman, Taylor Rayne, and Carsten Eickhoff. 2022.
\newblock \href {https://aclanthology.org/2022.findings-acl.262} {{I}so{S}core:
  Measuring the uniformity of embedding space utilization}.
\newblock In \emph{Findings of the Association for Computational Linguistics:
  ACL 2022}, pages 3325--3339, Dublin, Ireland. Association for Computational
  Linguistics.

\bibitem[{Sch{\"o}nemann(1966)}]{schonemann1966generalized}
Peter~H Sch{\"o}nemann. 1966.
\newblock A generalized solution of the orthogonal procrustes problem.
\newblock \emph{Psychometrika}, 31(1):1--10.

\bibitem[{S{\o}gaard et~al.(2018)S{\o}gaard, Ruder, and
  Vuli{\'c}}]{sogaard-etal-2018-limitations}
Anders S{\o}gaard, Sebastian Ruder, and Ivan Vuli{\'c}. 2018.
\newblock \href {https://doi.org/10.18653/v1/P18-1072} {On the limitations of
  unsupervised bilingual dictionary induction}.
\newblock In \emph{Proceedings of the 56th Annual Meeting of the Association
  for Computational Linguistics (Volume 1: Long Papers)}, pages 778--788,
  Melbourne, Australia. Association for Computational Linguistics.

\bibitem[{Su et~al.(2021)Su, Cao, Liu, and Ou}]{bert_whitening}
Jianlin Su, Jiarun Cao, Weijie Liu, and Yangyiwen Ou. 2021.
\newblock \href {http://arxiv.org/abs/2103.15316} {Whitening sentence
  representations for better semantics and faster retrieval}.
\newblock \emph{CoRR}, abs/2103.15316.

\bibitem[{Vuli{\'c} et~al.(2019)Vuli{\'c}, Glava{\v{s}}, Reichart, and
  Korhonen}]{vulic-etal-2019-really}
Ivan Vuli{\'c}, Goran Glava{\v{s}}, Roi Reichart, and Anna Korhonen. 2019.
\newblock \href {https://doi.org/10.18653/v1/D19-1449} {Do we really need fully
  unsupervised cross-lingual embeddings?}
\newblock In \emph{Proceedings of the 2019 Conference on Empirical Methods in
  Natural Language Processing and the 9th International Joint Conference on
  Natural Language Processing (EMNLP-IJCNLP)}, pages 4407--4418, Hong Kong,
  China. Association for Computational Linguistics.

\bibitem[{Vuli{\'c} et~al.(2020)Vuli{\'c}, Ruder, and
  S{\o}gaard}]{vulic-etal-2020-good}
Ivan Vuli{\'c}, Sebastian Ruder, and Anders S{\o}gaard. 2020.
\newblock \href {https://doi.org/10.18653/v1/2020.emnlp-main.257} {Are all good
  word vector spaces isomorphic?}
\newblock In \emph{Proceedings of the 2020 Conference on Empirical Methods in
  Natural Language Processing (EMNLP)}, pages 3178--3192, Online. Association
  for Computational Linguistics.

\bibitem[{Wang et~al.(2019)Wang, Che, Guo, Liu, and Liu}]{wang-etal-2019-cross}
Yuxuan Wang, Wanxiang Che, Jiang Guo, Yijia Liu, and Ting Liu. 2019.
\newblock \href {https://doi.org/10.18653/v1/D19-1575} {Cross-lingual {BERT}
  transformation for zero-shot dependency parsing}.
\newblock In \emph{Proceedings of the 2019 Conference on Empirical Methods in
  Natural Language Processing and the 9th International Joint Conference on
  Natural Language Processing (EMNLP-IJCNLP)}, pages 5721--5727, Hong Kong,
  China. Association for Computational Linguistics.

\bibitem[{Wang et~al.(2020)Wang, Xie, Xu, Yang, Neubig, and
  Carbonell}]{Wang*2020Cross-lingual}
Zirui Wang, Jiateng Xie, Ruochen Xu, Yiming Yang, Graham Neubig, and Jaime~G.
  Carbonell. 2020.
\newblock \href {https://openreview.net/forum?id=S1l-C0NtwS} {Cross-lingual
  alignment vs joint training: A comparative study and a simple unified
  framework}.
\newblock In \emph{International Conference on Learning Representations}.

\bibitem[{Wu and Dredze(2019)}]{wu-dredze-2019-beto}
Shijie Wu and Mark Dredze. 2019.
\newblock \href {https://doi.org/10.18653/v1/D19-1077} {Beto, bentz, becas: The
  surprising cross-lingual effectiveness of {BERT}}.
\newblock In \emph{Proceedings of the 2019 Conference on Empirical Methods in
  Natural Language Processing and the 9th International Joint Conference on
  Natural Language Processing (EMNLP-IJCNLP)}, pages 833--844, Hong Kong,
  China. Association for Computational Linguistics.

\bibitem[{Wu and Dredze(2020)}]{wu-dredze-2020-explicit}
Shijie Wu and Mark Dredze. 2020.
\newblock \href {https://doi.org/10.18653/v1/2020.emnlp-main.362} {Do explicit
  alignments robustly improve multilingual encoders?}
\newblock In \emph{Proceedings of the 2020 Conference on Empirical Methods in
  Natural Language Processing (EMNLP)}, pages 4471--4482, Online. Association
  for Computational Linguistics.

\bibitem[{Zhang et~al.(2022)Zhang, Buntine, and
  Shareghi}]{zhang-etal-2022-effect}
Lan Zhang, Wray Buntine, and Ehsan Shareghi. 2022.
\newblock \href {https://aclanthology.org/2022.acl-short.78} {On the effect of
  isotropy on {VAE} representations of text}.
\newblock In \emph{Proceedings of the 60th Annual Meeting of the Association
  for Computational Linguistics (Volume 2: Short Papers)}, pages 694--701,
  Dublin, Ireland. Association for Computational Linguistics.

\bibitem[{Zhang et~al.(2017)Zhang, Liu, Luan, and Sun}]{zhang-etal-2017-earth}
Meng Zhang, Yang Liu, Huanbo Luan, and Maosong Sun. 2017.
\newblock \href {https://doi.org/10.18653/v1/D17-1207} {Earth mover{'}s
  distance minimization for unsupervised bilingual lexicon induction}.
\newblock In \emph{Proceedings of the 2017 Conference on Empirical Methods in
  Natural Language Processing}, pages 1934--1945, Copenhagen, Denmark.
  Association for Computational Linguistics.

\bibitem[{Zhang et~al.(2019)Zhang, Xu, Kawarabayashi, Jegelka, and
  Boyd-Graber}]{zhang-etal-2019-girls}
Mozhi Zhang, Keyulu Xu, Ken-ichi Kawarabayashi, Stefanie Jegelka, and Jordan
  Boyd-Graber. 2019.
\newblock \href {https://doi.org/10.18653/v1/P19-1307} {Are girls neko or
  sh{\=o}jo? cross-lingual alignment of non-isomorphic embeddings with
  iterative normalization}.
\newblock In \emph{Proceedings of the 57th Annual Meeting of the Association
  for Computational Linguistics}, pages 3180--3189, Florence, Italy.
  Association for Computational Linguistics.

\end{thebibliography}
\bibliographystyle{acl_natbib}

\clearpage
\appendix

\setcounter{table}{0}
\renewcommand{\thetable}{A\arabic{table}}

\section{Appendix}
We wrote the proofs in Sections \ref{sec:app-white} and \ref{sec:app-ortho} in the process of understanding the advanced mapping procedure of VecMap.  The interested reader may find them elucidating. 

\label{sec:appendix}
\counterwithin{figure}{section}
\setcounter{figure}{0}  
\counterwithin{table}{section}
\setcounter{table}{0}  

\subsection{Whitening Transformations}
\label{sec:app-white}

\paragraph{Estimating the covariance matrix.}  Let $X \in \mathbb{R}^{n \times d}$ be a data matrix of $n$ samples of $d$ dimensions. We mean-center $X$ dimension-wise, then recall that the below is an unbiased estimator over $n$ samples of the covariance matrix of $X$. 

$$\Sigma_X = \frac{1}{(n-1)}{X}^T{X}$$

\paragraph{The whitening matrix.} 
The goal of a whitening transformation is to transform $X$ into a data matrix $Z \in \mathbb{R}^{n \times d}$ such that all dimensions are uncorrelated and have unit variance, meaning that the covariance matrix $\Sigma_Z = \frac{1}{(n-1)}{Z}^T{Z} = I$.  Doing so ensures that each basis vector in the space of $Z$ is minimally correlated with all others, and therefore retains maximal explanatory power.  We may state the whitening objective as finding $W$ such that $Z = XW^T$ and $\Sigma_Z = I$. Let $c = \frac{1}{n-1}$, so that $\Sigma_Z = c{Z}^T{Z}$.  

We claim that for $W$, if $W^TW = \Sigma_X^{-1}$, we obtain a valid whitening matrix:
\begin{align*}
    cZ^TZW &= c(XW^T)^TXW^TW \\
    & = cWX^TXW^TW \\
    &= W\Sigma_X W^TW \\
    &= W\Sigma_X \Sigma_X^{-1} \\
    &= W
\end{align*}
Therefore, since $cZ^T ZW = W$, $cZ^TZ = I$. 
There are infinite solutions for $W^TW = 
\Sigma_X^{-1}$.
A common choice called ``ZCA Whitening'' selects $W_{ZCA} = \Sigma_X^{-\frac{1}{2}}$. See \citet{kessy2018optimal} for a similar derivation and head-to-head comparison of 5 whitening methods. 

\paragraph{VecMap's whitening is proportional to ZCA.} 
VecMap's whitening chooses $W_{vecmap} = V S^{-1} V^T$, where $USV^T = \text{SVD} (X)$. We show that this choice is proportional to ZCA Whitening, and furthermore that it is a valid whitening matrix.

First, we show that $W_{vecmap}$ is proportional to $W_{ZCA} = \Sigma_X^{-\frac{1}{2}}$. 
Note that $S^T = S$ and that $S^{-1}$ can be written as $\frac{1}{S}$ because $S$ is diagonal. 
\begin{align*}
    \Sigma_X &= c(USV^T)^T USV^T \\
    &= cVS^TU^TUSV^T \\
    & = cVS^2V^T \\
    \implies \Sigma_X^{\frac{1}{2}} &= \sqrt{c}VSV^T \\
    \implies W_{ZCA} = \Sigma_X^{-\frac{1}{2}} &= \frac{1}{\sqrt{c}}VS^{-1}V^T \\
    &\propto W_{vecmap}
\end{align*}

\paragraph{$\bf{W_{vecmap}}$ is a valid whitening matrix.}
Let $Z = XW_{vecmap}^T$. We claim that if $Z = XW_{vecmap}^T$, then $\Sigma_Z = cZ^TZ = I$. We assume $X$ is unit-normalized and mean-centered.\footnote{Later work on VecMap re-normalizes the embeddings after mean-centering, and is the version used in this work.}
\begin{align*}
    cZ^TZ &= c(XW^T)^TXW^T = cWX^TXW^T \\
    &= cVS^{-1}V^T X^TX (VS^{-1}V^T)^T \\
    &= cVS^{-1}V^TX^TXVS^{-1}V^T \\
    &= cVS^{-1}V^T(USV^T)^TUSV^TVS^{-1}V^T \\
    &= cVS^{-1}V^TVSU^TUSV^TVS^{-1}V^T \\
    &= cI
\end{align*}

\subsection{Derivation of Orthogonal Mapping Variant Before Dictionary Induction in VecMap}
\label{sec:app-ortho}
The orthogonal mapping performed at the final step of VecMap is a variant of the orthogonal Procrustes problem.  VecMap finds the matrices $W_x$ and $W_z$ such that $XW_x$ and $ZW_z$ are maximally close to each other. This is equivalent to minimizing $||ZWz - XWx||$.  The solution is $W_z = U$ and $Wx = V$ where $USV^T = \text{SVD}(X^TZ)$.  Observe the derivation below, where we use the Frobenius norm/inner product:
\begin{align*}
&\underset{W_x, W_x}{\arg\min} || ZW_z - XW_x || \\
= &\underset{W_x, W_x}{\arg\min}\langle ZW_z - XW_x , ZW_z - XW_x \rangle\\
= &\underset{W_x, W_x}{\arg\min}\langle ZW_z, ZW_z\rangle - 2\langle XW_x, ZW_z\rangle \\ &\;\;\;\;\;\;\;\;\;\;\;\;\;\;\;\;\;\;\;\;\;\;\;\;\;\;\;\;+\langle XW_x, XW_x\rangle \\
= &\underset{W_x, W_x}{\arg\min} || Z ||^2 - 2\langle XW_x, ZW_z \rangle + || X ||^2 \\
= &\underset{W_x, W_x}{\arg\min} -2\langle XW_x, ZW_z \rangle \\
= &\underset{W_x, W_x}{\arg\max} \langle XW_x, ZW_z \rangle \\
= &\underset{W_x, W_x}{\arg\max} \text{ Trace}(W_z^TZ^TXW_x)
\end{align*}
Given the cyclic property of the trace, the objective is equivalent to maximizing $\text{ Trace}(XW_xW_z^TZ^T)$, as stated by \citet{artetxe2018generalizing}.

\subsection{RSIM vs. P@1: Pearson's Correlation}
\begin{figure}[!h]
\centering
\includegraphics[width=1\linewidth]{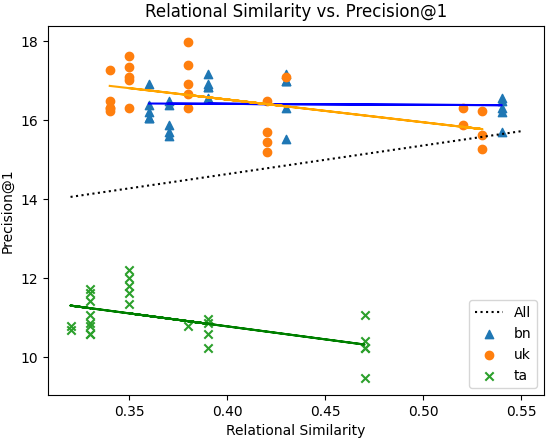}
\caption{Relational similarity vs. P@1 for all supervised \textit{IsoVec} experiments (See Section \ref{sec:effect-on-iso}).}
\label{fig:rsim-v-p1}
\end{figure}

Figure \ref{fig:rsim-v-p1} shows relational similarity score vs. P@1 over all language pairs described in Section \ref{sec:effect-on-iso}. We observe how it is possible to have within-language negative correlations but a positive overall correlation. 

\subsection{Effect on Isomorphism, After Mapping}
\label{sec:app-iso-mapped}

\begin{table*}[htb]
\centering  \setlength\tabcolsep{3.5pt}\footnotesize
\begin{tabular}{l|ccc|ccc|ccc}
\toprule
& \multicolumn{3}{c}{\underline{Relational Sim. $\uparrow$}} & \multicolumn{3}{c}{\underline{GH Distance $\downarrow$}} & \multicolumn{3}{c}{\underline{Eigenvector Sim $\downarrow$}} \\
& bn & uk & ta & bn & uk & ta & bn & uk & ta \\
\midrule
\hspace{1mm}\textit{Baseline} & \textit{0.43} & \textit{0.35} & \textit{0.38} & \textit{0.41} & \textit{1.02} & \textit{0.42} & \textit{92.8} & \textit{113.8} & \textit{133.3} \\
\hspace{1mm}L2 & 0.47 & 0.41 & 0.41 & 0.36 & 0.91 & 0.38 & 97.8 & 145.3 & 98.3 \\
\hspace{1mm}Proc-L2 & 0.51 & 0.48 & 0.45 & 0.35 & 0.81 & 0.37 & 87.4 & 148.2 & 96.8 \\
\hspace{5mm}+\textit{Init} & 0.48 & 0.44 & 0.43 & 0.37 & 0.68 & \textbf{0.35} & 97.4 & 155.4 & 99.4 \\
\hspace{1mm}RSIM & \textbf{0.65} & \textbf{0.62} & \textbf{0.59} & \textbf{0.34} & 0.71 & 0.40 & \textbf{43.2} & \textbf{108.8} & \textbf{69.2} \\
\hspace{5mm}+\textit{Init}  & 0.48 & 0.45 & 0.43 & 0.37 & 0.62 & 0.37 & 95.5 & 126.4 & 96.6 \\
\midrule
\hspace{1mm}RSIM-U & 0.43 & 0.35 & 0.37 & 0.37 & \textbf{0.59} & 0.44 & 94.6 & 121.9 & 135.5 \\
\hspace{1mm}EVS-U  & 0.43 & 0.35 & 0.37 & 0.48 & 0.64 & 0.49 & 83.1 & 109.4 & 110.7 \\
\bottomrule
\end{tabular}
\caption{Isomorphism scores of source vs. reference embedding space after training \textit{and} mapping with VecMap. All experiments are mapped in semi-supervised mode.  Results averaged over 5 runs.  Best score is \textbf{bold}. }
\label{app:final-iso-scores-mapped}
\end{table*}

Comparing with Table \ref{tab:final-iso-scores} in the main body of this paper, Table \ref{app:final-iso-scores-mapped} shows average isomorphism scores of source vs. reference embedding spaces \textit{after} mapping with VecMap in semi-supervised mode.\footnote{Though we map the Baseline in supervised mode and unsupervised methods in unsupervised mode in the main body of the paper, we map everything in semi-supervised mode here for comparability.}  \textbf{RSIM} best improves relational similarity and eigenvector similarity. GH distance improves for supervised methods and \textbf{RSIM-U}.

Table \ref{tab:final-iso-scores} measures the unperturbed geometry of the space after applying \textit{IsoVec}.
Importantly, RelSim, GH, and EigSim \textit{do not require mapping} for measurement, as they are invariant to transformation (RelSim and EigSim measure nearest-neighbor graphs, and GH measures nearest neighbor after optimal isometric transform).  Table \ref{app:final-iso-scores-mapped}, measures a geometry that may have been perturbed due to VecMap operations such as whitening/dewhitening. 
While isomorphism scores over the mapped spaces appear more consistent in terms of internal patterns, Table \ref{tab:final-iso-scores} measures isomorphism induced directly by \textit{IsoVec}, whereas Table \ref{app:final-iso-scores-mapped} may be influenced by VecMap as well.  

\begin{table}[htb]
\centering  \setlength\tabcolsep{3.5pt}\footnotesize
\begin{tabular}{l|c|r|rrr}
\toprule
 & & Overall & bn-en & uk-en & ta-en \\
 \midrule
P@1 \hspace{1mm} vs. RelSim & (+) & 0.13 & {\color[HTML]{FF0000} -0.16} & {\color[HTML]{FF0000} -0.49} & {\color[HTML]{FF0000} -0.59} \\
\hspace{8mm} vs. GH & (-) & {\color[HTML]{FE0000} 0.39} & {\color[HTML]{C0C0C0} 0.02} & {\color[HTML]{000000} -0.42} & -0.79 \\
\hspace{8mm} vs. EigSim & (-) & {\color[HTML]{FF0000} 0.31} & {\color[HTML]{FF0000} 0.22} & {\color[HTML]{FF0000} 0.15} & -0.29 \\
\midrule
RelSim \hspace{0.05mm} vs. GH  & (-) & -0.12 & -0.54 & {\color[HTML]{000000} -0.20} & {\color[HTML]{FF0000} 0.51} \\
\hspace{8mm} vs. EigSim & (-) & -0.54 & -0.88 & -0.57 & -0.10 \\
GH  \hspace{2mm} vs. EigSim & (+) & 0.72 & 0.57 & 0.23 & 0.72 \\
\bottomrule
\end{tabular}
\caption{Pearson's Correlations: BLI performance (P@1) vs. relational similarity, Gromov-Hausdorff Distance, and eigenvector similarity score of trained source embeddings vs. English reference space, \textit{after VecMap mapping in semi-supervised mode}, and isomorphism metrics correlated amongst themselves. Computed over supervised \textit{IsoVec} experiments (ex. P@1 vs. RelSim for bn-en is P@1 from all runs of {L2}, {Proc-L2}, {Proc-L2+I}, {RSIM}, {RSIM-Init} with RelSim/GH/EigSim for each.) Directly comparable to Table \ref{tab:corrs} (right).}
\label{tab:app-corrs-mapped-sup}
\end{table}

\begin{table}[htb]
\centering  \setlength\tabcolsep{3.5pt}\footnotesize
\begin{tabular}{l|c|r|rrr}
\toprule
 & & Overall & bn-en & uk-en & ta-en \\
 \midrule
P@1 \hspace{1mm} vs. RelSim & (+) & 0.29 & 0.44 & 0.39 & 0.14 \\
\hspace{8mm} vs. GH & (-) & {\color[HTML]{FE0000} 0.34} & -0.26 & {\color[HTML]{FF0000} 0.19} & -0.78 \\
\hspace{8mm} vs. EigSim & (-) & {\color[HTML]{FF0000} 0.14} & {\color[HTML]{C0C0C0} 0.00} & {\color[HTML]{FF0000} 0.42} & -0.29 \\
\midrule
RelSim \hspace{0.05mm} vs. GH  & (-) & -0.15 & -0.33 & {\color[HTML]{FF0000} 0.21} & -0.34 \\
\hspace{8mm} vs. EigSim & (-) & -0.48 & -0.75 & -0.06 & -0.69 \\
GH  \hspace{2mm} vs. EigSim & (+) & 0.63 & 0.20 & 0.33 & 0.31 \\
\bottomrule
\end{tabular}
\caption{Same as Table \ref{tab:app-corrs-mapped-sup}, except computed over \textit{all} \textit{IsoVec} experiments ({L2}, {Proc-L2}, {Proc-L2+I}, {RSIM}, {RSIM-Init}, {RSIM-U}, {EVS-U}).}

\label{tab:app-corrs-mapped}
\end{table}

Table \ref{tab:app-corrs-mapped-sup} shows correlations after mapping over only the supervised \textit{IsoVec} methods.  Compared with Table \ref{tab:corrs} (right), we see that relationships between the isomorphism measures are more aligned with our expectations now, but still no isomorphism measure is consistently predictive of P@1 across languages.   Table \ref{tab:app-corrs-mapped} shows the same calculations over \textit{all} \textit{IsoVec} experiments mapped in semi-supervised mode, though unsupervised training with semi-supervised mapping probably would not be used in practice as it is here for \textbf{RSIM-U} and \textbf{EVS-U}. 

\end{document}